\documentclass[10pt,journal,compsoc]{IEEEtran}

\usepackage{CJKutf8}
\usepackage{soul}
\usepackage{url}
\usepackage[utf8]{inputenc}
\usepackage{caption}
\usepackage{graphicx}
\usepackage{amsmath}
\usepackage{amsthm}
\usepackage{booktabs}
\usepackage{algorithm}
\usepackage{algorithmic}
\usepackage{amsfonts}
\usepackage{color}
\usepackage{dblfloatfix}
\usepackage{balance}
\usepackage{cite} 
\usepackage{makecell}
\usepackage{hyperref}
\usepackage{nicematrix}

\usepackage{subcaption}
\usepackage{bm}
\usepackage{multirow}
\usepackage{enumitem}
\usepackage{colortbl}
\definecolor{mygray}{gray}{.9}
\definecolor{mypink}{rgb}{.99,.91,.95}
\definecolor{mycyan}{cmyk}{.3,0,0,0}

\newtheorem{definition}{Definition}

\newcommand{\eg}{{\it e.g.}}
\newcommand{\etal}{{\it et al.}}
\newcommand{\ie}{{\it i.e.}}
\newcommand{\iid}{{\it i.i.d.}}
\newcommand{\etc}{{\it etc.}}

\newcommand{\wrt}{w.r.t. }
\newcommand{\stitle}[1]{\vspace{1mm} \noindent {\bf #1}}
\newcommand{\ititle}[1]{\vspace{1mm} \noindent {\it #1}}

\newcommand{\name}{ILoGs{}}

\newcommand{\bG}{\ensuremath{\mathcal{G}}}
\newcommand{\bC}{\ensuremath{\mathcal{C}}}
\newcommand{\bV}{\ensuremath{\mathcal{V}}}
\newcommand{\bE}{\ensuremath{\mathcal{E}}}
\newcommand{\bD}{\ensuremath{\mathcal{D}}}

\newcommand{\bN}{\ensuremath{\mathcal{N}}}

\newcommand{\bT}{\ensuremath{\mathcal{T}}}
\newcommand{\bR}{\ensuremath{\mathcal{R}}}
\newcommand{\bM}{\ensuremath{\mathcal{M}}}

\newcommand{\eat}[1]{}

\newcommand\del[1]{{#1}}

\let\vec\mathbf

\ifCLASSINFOpdf
\else
\fi
\begin{document}
%
\title{A Survey of Imbalanced Learning on Graphs: \\ Problems, Techniques, and Future Directions}

\author{Zemin Liu,
        Yuan Li,
        Nan Chen,
        Qian Wang,
        Bryan Hooi,
        Bingsheng He
\thanks{We have curated a collection of awesome literature on imbalanced learning on graphs on GitHub: \url{https://github.com/Xtra-Computing/Awesome-Literature-ILoGs}.}
\IEEEcompsocitemizethanks{\IEEEcompsocthanksitem  Z.~Liu, Y.~Li, N.~Chen, Q.~Wang, B.~Hooi, and B.~He are with School of Computing, National University of Singapore.
E-mail: \{zeminliu, li.yuan, nanchen, qiansoc, c\}@nus.edu.sg, \{hebs\}@comp.nus.edu.sg.
}
}

%
%

\markboth{Journal of \LaTeX\ Class Files,~Vol.~14, No.~8, August~2015}%
{Shell \MakeLowercase{\textit{et al.}}: Bare Demo of IEEEtran.cls for Computer Society Journals}
%



\IEEEtitleabstractindextext{%
\begin{abstract}
Graphs represent interconnected structures prevalent in a myriad of real-world scenarios. Effective graph analytics, such as graph learning methods, enables users to gain profound insights from graph data, underpinning various tasks including node classification and link prediction. However, these methods often suffer from data imbalance, a common issue in graph data where certain segments possess abundant data while others are scarce, thereby leading to biased learning outcomes. This necessitates the emerging field of imbalanced learning on graphs, which aims to correct these data distribution skews for more accurate and representative learning outcomes.
In this survey, we embark on a comprehensive review of the literature on imbalanced learning on graphs. We begin by providing a definitive understanding of the concept and related terminologies, establishing a strong foundational understanding for readers. 
Following this, we propose two comprehensive taxonomies: (1) the \emph{problem taxonomy}, which describes the forms of imbalance we consider, the associated tasks, and potential solutions; (2) the \emph{technique taxonomy}, which details key strategies for addressing these imbalances, and aids readers in their method selection process.
Finally, we suggest prospective future directions for both problems and techniques within the sphere of imbalanced learning on graphs, fostering further innovation in this critical area. 
\end{abstract}

}

\maketitle

\IEEEdisplaynontitleabstractindextext

%
\IEEEpeerreviewmaketitle

\IEEEraisesectionheading{\section{Introduction}\label{sec.introduction}}

\IEEEPARstart{G}{raphs}, or networks, refer to interconnected structures that are commonly found in real-world scenarios, where entities often interact with each other.
Graphs are ubiquitous in various domains, such as social networks on platforms like Facebook, citation networks on DBLP, e-commerce networks on Amazon, \etc\
The prevalence of graph structures has sparked significant interest in graph analytics, which aims to leverage the inherent information in graphs for downstream tasks such as node classification, link prediction, and graph classification. 

Early studies in graph analytics often rely on conventional techniques like feature engineering \cite{page1999pagerank,jeh2002simrank,backstrom2011supervised}, which can be computationally expensive and require significant effort. 
However, with the advent of graph representation learning \cite{cai2018comprehensive,wu2020comprehensive}, new opportunities have emerged for graph analytics. Graph representation learning aims to embed the structures of graphs (\eg, nodes, edges, or graphs) into a low-dimensional space while preserving their structural information.
Prior graph embedding approaches \cite{cai2018comprehensive} such as Deepwalk \cite{perozzi2014deepwalk}, LINE \cite{tang2015line}, and node2vec \cite{grover2016node2vec} rely on contextual connectivity between nodes to capture their proximity for node representation learning. 
Recently, more attention has shifted to graph neural networks (GNNs) \cite{kipf2016semi,hamilton2017inductive,velivckovic2017graph,xu2018powerful}, a family of graph representation learning approaches that capitalize on neighborhood aggregation. They conduct neighborhood aggregation to recursively pass and receive messages along the edges in an end-to-end manner, to effectively encode the graph structure. Consequently, GNNs have achieved state-of-the-art performance in many downstream tasks.

\stitle{Imbalance phenomenon.}
While these graph representation learning approaches can be effective, like many machine learning models, they often require a large amount of labeled data for training. Real-world data, however, frequently display imbalanced distributions, where some segments have an abundance of data and others do not. For instance, in a classification task, the distribution of labeled data---such as images or documents---may skew towards certain classes, resulting in imbalanced label distribution. This data imbalance can notably impact the training process. Specifically, the model tends to be well-trained on \emph{high-resource} groups that have ample data, while underperforming on \emph{low-resource} groups that are data-limited, leading to unclear class boundaries \cite{buda2018systematic,cui2019class,Kang2020Decoupling}. As a result, the model performance is usually satisfactory within high-resource groups, but degrades within low-resource groups \cite{chawla2002smote,zhang2023deep,liu2020towards}. Thus, severe challenges of imbalanced learning have garnered significant attention in the literature and addressing those challenges is of paramount importance \cite{he2009learning,krawczyk2016learning,haixiang2017learning,johnson2019survey,zhang2023deep}. 

\begin{figure*}[t]
\centering
\includegraphics[width=1\linewidth]{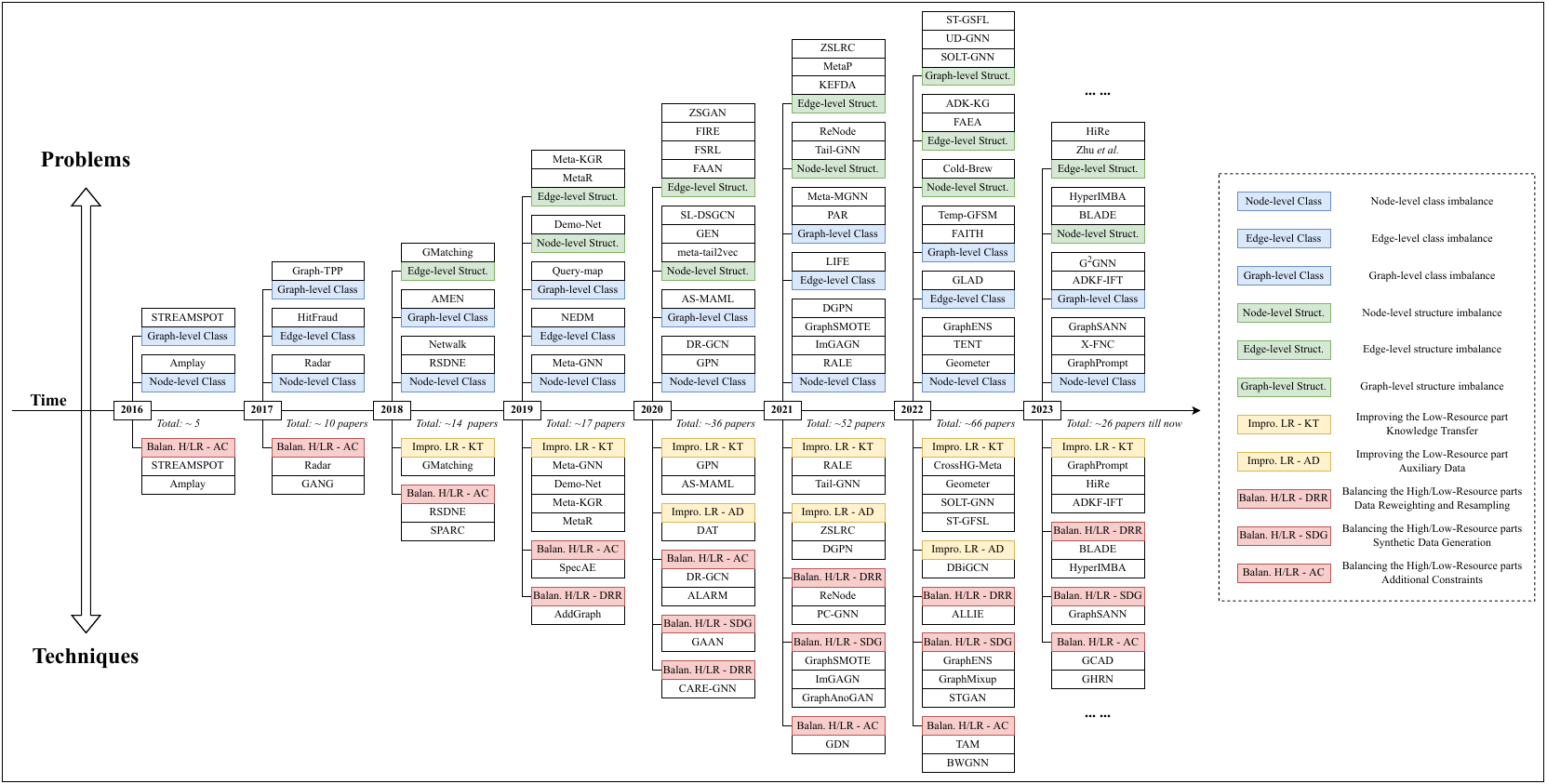} 
\caption{This timeline showcases the representative research literature on \name\ in relation to our taxonomies, \ie, Problems and Techniques. The total number of research works for each year is also displayed, illustrating an upward trend.}
\label{fig.intro-timeline}
\end{figure*}

\stitle{Imbalanced learning on graphs.}
To tackle the imbalance issues, various approaches have been proposed in the fields of vision and language \cite{huang2016learning,li2020dice}. 
However, graph data differs from them in the fact that samples are inherently non-\iid\ and multifarious, with diverse structural aspects (\eg, node degrees). Directly applying these approaches to address imbalance issues on graphs may not be feasible \cite{zhao2021graphsmote,yao2020graph,liu2021tail}.
Thus, graph learning faces the challenge of new imbalance types beyond those found in these settings.

The considerable impact of imbalance issues on the performance of graph-based tasks has recently attracted significant research interest \cite{ma2021comprehensive,mandal2022metalearning, zhang2022few}, as illustrated in Fig.~\ref{fig.intro-timeline}. The annually increasing volume of literature mirrors the escalating importance and substantial impact of addressing challenges in imbalanced learning within the realm of graph learning tasks.
These research efforts endeavor to address a variety of real-world applications \cite{yao2020graph,zhu2023few,wang2022imbalanced,liu2020towards,xiong2018one,liu2022size}, formalized into several distinct research \emph{problems}. Each problem, characterized by its unique features, calls for the development of specialized \emph{techniques} tailored to effectively address the imbalance issues specific to each scenario.
However, the diversity of problems and techniques has resulted in a scattered landscape of imbalanced learning on graphs, lacking a comprehensive framework to identify their commonalities and disparities. 

\begin{figure*}[t]
\centering
\includegraphics[width=1\linewidth]{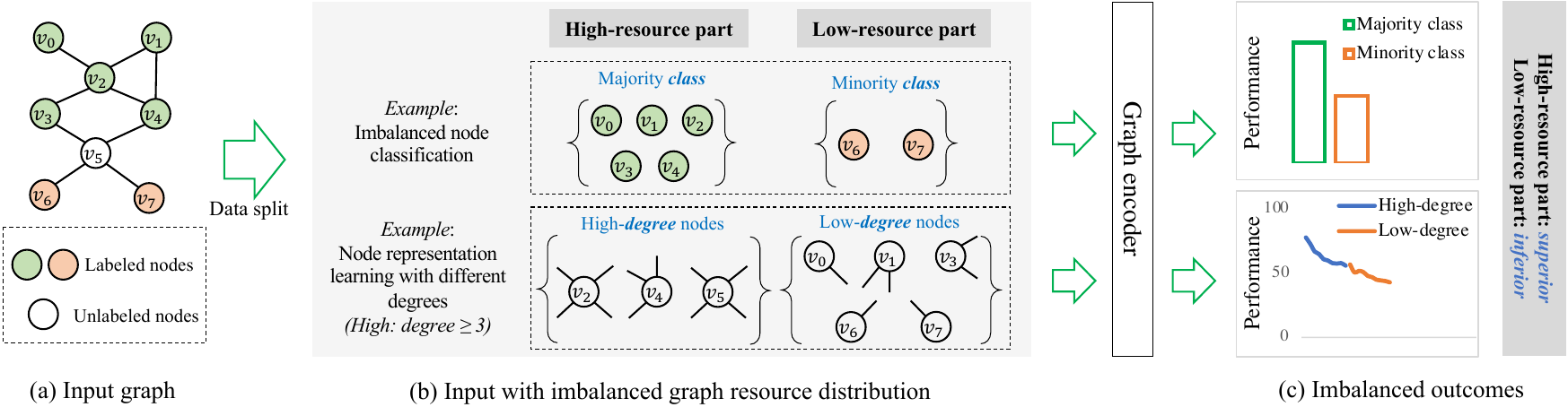} 
\caption{Imbalanced learning on graphs: imbalanced graph resource distribution results in imbalanced outcomes.}
\label{fig.intro-imba-problem}
\end{figure*}

Therefore, in this survey, we focus on \textbf{I}mbalanced \textbf{L}earning \textbf{o}n \textbf{G}raph\textbf{s} (\textbf{\name}) to bridge this gap by reviewing and summarizing the \emph{problems} and \emph{techniques} in the context of addressing imbalance issues on graphs. The essence of \name\ lies in the observation that graph learning models with imbalanced input typically exhibit varying performance across groups with different levels of graph resource abundance \cite{shi2020multi,liu2021pick,wang2022imbalanced,liu2020towards,liu2022size}. 
More precisely, as shown in Fig.~\ref{fig.intro-imba-problem}, given the input graph(s), graph data is often divided into multiple segments, creating an imbalanced distribution of graph resources. This phenomenon can be observed in various tasks, such as imbalanced node classification \cite{zhao2021graphsmote,qu2021imgagn} and node representation learning with different degrees \cite{liu2020towards,liu2021tail}, as depicted in Fig.~\ref{fig.intro-imba-problem}(b). This uneven distribution often leads to imbalanced outcomes: graph models typically perform well on high-resource parts while marginalizing the low-resource parts, ultimately resulting in disparities in performance across parts, as shown in Fig.~\ref{fig.intro-imba-problem}(c).

However, the multifaceted nature of imbalanced learning on graphs presents considerable complexity in its investigation.
On one hand, graph structures spawn a wide array of \emph{problems} characterized by varying forms of imbalance, tasks, and solutions. This diversity presents a challenge in unifying these elements. Therefore, creating an organized taxonomy to categorize these imbalanced learning problems on graphs is a significant task. Additionally, a detailed categorization will benefit future research by identifying unexplored areas.
On the other hand, the numerous imbalance problems on graphs yield diverse solutions, and some tasks demand specific \emph{techniques} due to their unique attributes. This creates a complex landscape of solutions, making research in this area challenging. Therefore, classifying literature from a technical perspective is crucial. Moreover, this classification can also aid readers in selecting appropriate techniques to handle their specific graph imbalance issues.

To resolve the complexity, we categorize the literature from the perspectives of problems and techniques to provide a comprehensive overview.
Primarily, for the taxonomy of \emph{problems}, we classify the literature based on class imbalance and structure imbalance, both stemming from imbalanced input. We further distill this into more specific categories: node-, edge-, and graph-level imbalance, offering a comprehensive understanding of graph imbalance, as depicted in Fig.~\ref{fig.taxonomy-problems}.
Besides, to provide a clearer insight into these imbalance issues, we delve deeper by detailing and contrasting their types, settings, and information abundance in Table~\ref{table.graph-imba-issues}.
Additionally, for the taxonomy of \emph{techniques}, we classify literature based on imbalance types and the associated strategies to mitigate it, as summarized in Fig.~\ref{fig.taxonomy-techniques}. This is because, specific types of imbalance may require particular techniques to effectively address the associated issues, and the techniques employed may vary across different imbalance types.

Using these taxonomies allows us to capture commonalities and disparities across imbalanced graph learning literature. These hierarchical taxonomies group related studies based on shared features, such as imbalance type, imbalance level (\ie, node, edge, and graph), and techniques. Simultaneously, the individual characteristics of each research task or technique are highlighted, underscoring the unique aspects that set them apart from each other. This approach provides a holistic view of the field, shedding light on both shared trends and unique advances.
A more in-depth exploration of these taxonomies will be provided in Section~\ref{sec.overview}.

Building on the taxonomies of problems and techniques, we further delineate promising directions for future research in the realm of imbalanced learning on graphs. Specifically, when considering future directions concerning \emph{problems}, we dissect potential research challenges through the lenses of class imbalance and structure imbalance. As for future directions concerning \emph{techniques}, we contemplate innovative solutions that could fuel the growth of this field.

Note that, given the sheer volume of research papers tackling this intricate issue, our survey does not delve into the nuances of each study. Instead, our primary objective is to present a holistic, structured overview in line with our established taxonomies. This approach ensures we offer a broad perspective on the domain and spotlight the emergent frontiers in imbalanced learning on graphs.

\stitle{Relationship with existing surveys.}
Imbalanced learning has been the subject of several surveys in the past decade, covering various areas from general imbalanced classification \cite{he2009learning, krawczyk2016learning, fernandez2018smote} to more specific tasks like anomaly detection \cite{chandola2009anomaly, chalapathy2019deep, pang2021deep}, few-shot learning \cite{wang2020generalizing, song2023comprehensive}, long-tailed distribution \cite{zhang2023deep}, \etc\
However, these surveys have primarily focused on imbalanced learning in a general context or in specific tasks, and lack comprehensive coverage of imbalanced learning on graphs.
In the domain of graph-related tasks, there are surveys that focus on specific tasks on graphs such as class-imbalanced learning \cite{ma2023class}, anomaly detection \cite{bhuyan2013network, akoglu2015graph, ahmed2016survey, ma2021comprehensive}, few-shot classification \cite{mandal2022metalearning, zhang2022few}, and fairness learning \cite{dong2023fairness}. Despite their relevance, these surveys concentrate on individual tasks and lack a comprehensive overview of imbalanced learning on graphs.

Our survey fills this gap by providing a holistic view of imbalanced learning on graphs, covering diverse tasks with a focus on both class imbalance and structure imbalance. In contrast to previous surveys \cite{bhuyan2013network, akoglu2015graph, ahmed2016survey, ma2021comprehensive, mandal2022metalearning, zhang2022few, dong2023fairness}, our work elucidates the shared traits and unique characteristics of these tasks, offering fresh insights into their commonalities and differences within the sphere of imbalanced learning on graphs, based on the two taxonomies. 
Furthermore, we also incorporate fairness learning on graphs (see Section~\ref{sec.fairness-learning} for more details) as a specific type of imbalanced learning. Unlike other forms of imbalanced learning, fairness learning focuses on mitigating the potential bias and discrimination that may exist in the predictions made by the models, rather than solely on improving performance on imbalanced data.

\stitle{Contributions.}
To summarize, our survey makes the following major contributions.
\begin{itemize}[leftmargin=*]
\setlength\itemsep{1mm}
\item We present the first comprehensive survey that broadly encompasses the field of imbalanced learning on graphs. Given the importance of this research area and the increasing number of papers exploring it, our survey serves as an invaluable resource for both researchers and practitioners.
\item To provide a comprehensive and structured overview of the field, we introduce two novel taxonomies: problems and techniques. These taxonomies are designed to facilitate a thorough understanding of the existing literature, and provide a clear picture of the commonalities and distinctions through structured categorizations.
\item We identify potential future research directions in the field of imbalanced learning on graphs, providing insights and guidance for those interested in advancing the state-of-the-art in this fast-paced field.
\end{itemize}

\section{Background} \label{sec.background}

We present the main terminologies and definitions in this survey, for graphs and imbalanced learning, respectively.

\subsection{Graphs and Graph Representation Learning}

\subsubsection{Graph Formalization}

\stitle{Graph.} A \textbf{graph} can be represented as $G=\{\bV, \bE, \vec{X}_v, \vec{X}_e, \phi, \varphi, \bT, \bR\}$, where $\bV$ is the set of nodes, $\bE$ is the set of edges, $\vec{X}_v\in\mathbb{R}^{|\bV|\times d_{\vec{X}_v}}$ and $\vec{X}_e\in\mathbb{R}^{|\bV|\times d_{\vec{X}_e}}$ are the feature matrices of nodes and edges, respectively, and $\bT$ and $\bR$ are the sets of node types and edge types. For simplicity, we utilize $\vec{x}_v\in\mathbb{R}^{d_{\vec{X}_v}}$ and $\vec{x}_e\in\mathbb{R}^{d_{\vec{X}_e}}$ to denote the feature vectors of node $v$ and edge $e$, respectively. 
The function $\phi: \bV\rightarrow \bT$ maps a node $v\in \bV$ to its corresponding node type $\phi(v)$, while the function $\varphi: \bE\rightarrow \bR$ maps an edge $e_{\langle u,v\rangle}\in \bE$ to its corresponding edge type $\varphi(e_{\langle u,v\rangle})$.

The variety of node and edge types leads to the differentiation of graphs into various forms. 
A \textbf{homogeneous graph} is a graph without considering the node and edge types, \ie, $|\bT|=|\bR|=1$, and has been the subject of most research in graph analytics due to its simplicity.
A \textbf{heterogeneous graph} \cite{yang2020heterogeneous}, also called a Heterogeneous Information Network (HIN), is a graph that has different node or edge types, \ie, $|\bT|+|\bR|>2$, and preserves diverse semantics due to the presence of node and edge types.
In particular, heterogeneous graphs can be further classified into several categories, based on the diversity of heterogeneity. 
A \textbf{bipartite graph} requires that there only exist two types of nodes, with edges connecting nodes with different types, 
\ie, $|\bT|=2$ and $|\bR|=1$, and $\phi(u)\neq\phi(v)$ $\forall e_{\langle u,v\rangle}\in \bE$. Bipartite graphs are extensively investigated in the field of recommender systems \cite{zhang2019deep} by formalizing the user-item interactions into graph format.
Additionally, a \textbf{knowledge graph} (KG for short) \cite{ji2021survey} is a graph that consists of various types of edges between nodes, which are also known as \emph{relations} and \emph{entities}, respectively. KGs are prevalently employed in natural language processing to represent extracted knowledge from natural language.

For further information on these graph types, we refer readers to the relevant surveys mentioned above. 
As the main focus of this survey is on imbalanced learning on graphs, we do not extensively discuss or investigate the specific issues in these fields.

\subsubsection{Graph Representation Learning}

Graph representation learning \cite{cai2018comprehensive,wu2020comprehensive} is a powerful and efficient way for graph analytics. It offers a cost-effective alternative to traditional graph engineering techniques by mapping graph structures (\eg, nodes) into a low-dimensional space while preserving the structural information. 

Formally, let $\vec{h}_v\in\mathbb{R}^d$ denote the representation of node $v$. Node representation learning can be formalized as 
\begin{equation}
    \vec{h}_v = f(v,G;\theta_f),
\end{equation}
where $\theta_f$ represents the learnable parameters of the graph representation learning function $f(\cdot;\theta_f)$. 
Researchers can devise various objectives for the function $f(\cdot;\theta_f)$.

Recently, more attention has been shifted to GNNs \cite{wu2020comprehensive} which are based on the key operation of neighborhood aggregation. GNNs aggregate messages from neighboring nodes recursively to generate node representations, enabling end-to-end learning. Formally, in the $l$-th GNN layer, the representation of node $v$ can be calculated as 
\begin{equation}
    \vec{h}^l_v = \textsc{Aggr}(\vec{h}^{l-1}_v, \{\vec{h}^{l-1}_u: u\in\bN_v\}\mid \theta^l_g),
\end{equation}
where $\bN_v$ is the neighbors set of node $v$, $\textsc{Aggr}(\cdot)$ is the aggregation function, and $\theta^l_g$ is the parameters set in layer $l$. 
The aggregation function can take various forms, such as mean \cite{kipf2016semi}, sum \cite{xu2018powerful}, or attentive aggregation \cite{velivckovic2017graph}.
In the first layer, node representations can be initialized with node feature vectors, \ie, $\vec{h}^0_v=\vec{x}_v$. 
For simplicity, we directly use $\vec{h}_v$ to denote the output representation of node $v$.

Furthermore, graph-level representations \cite{ying2018hierarchical} are often needed for graph-level tasks, such as molecular classification \cite{guo2022graph}. They typically leverage learned substructure representations (\eg, nodes \cite{xu2018powerful} or edges \cite{yu2023learning}) for further aggregation to obtain graph representation. 
Given a graph $G$ and considering node representations as the foundation, graph representation learning can be formalized as
\begin{equation}
    \vec{h}_G = \textsc{ReadOut}(\{\vec{h}_v:v\in \bV\}; \theta_r),
\end{equation}
where $\textsc{ReadOut}(\cdot)$ is the readout function that aggregates node representations into the graph representation, parameterized by $\theta_r$.
In general, the $\textsc{ReadOut}$ function can primarily be implemented in two ways, \ie, global-pooling \cite{duvenaud2015convolutional,xu2018powerful} and hierarchical-pooling \cite{ying2018hierarchical,gao2019graph}.

\subsection{Imbalanced Learning on Graphs}

\stitle{Imbalanced learning.}
The conventional approaches for imbalanced learning primarily focus on developing learning algorithms to handle imbalanced classes \cite{he2009learning,krawczyk2016learning,fernandez2018smote}.

\begin{definition}[Conventional Imbalanced Learning]\label{def.imba-learn}
In the context of conventional imbalanced learning, consider a labeled data set $\bD=\{(x_i,y_i)\}^N_{i=1}$, which can be partitioned into $K$ classes (groups) such that $\bD=\bigcup_{1\leq j\leq K}\bG_j$ (given each group $\bG_j=\{(x_i,y_i):y_i=j\}$). There exists a notable imbalance in the number of labeled samples across these groups.
Under this setting, the imbalanced distribution of samples across groups would give rise to biases in the performance of a learning algorithm.
In particular, the low-resource groups, \ie, the classes with less labeled data, are usually marginalized by the learning model due to the domination of the high-resource groups, resulting in performance degradation for the former.
The goal of imbalanced learning is to develop a balanced model that can improve the performance of low-resource groups, potentially reaching levels comparable to those of high-resource groups.
\end{definition}

The degree of imbalance in a given dataset can be evaluated using the \emph{imbalance ratio}, a metric particularly pertinent in classification tasks \cite{zhang2023deep,zhao2021graphsmote}. Formally, for a labeled dataset encompassing $K$ classes (groups) $\bD=\bigcup_{1\leq i\leq K}\bG_i$, and with classes ordered by their cardinality in a descending manner (\ie, if $i<j$, then $|\bG_i|\ge |\bG_j|$, where $|\bG_i|$ denotes the size of group $\bG_i$), the imbalance ratio is defined as $|\bG_1|/|\bG_K|$. Essentially, the imbalance ratio quantifies the skewness of the distribution across classes in terms of their labeled data---a larger ratio signifies a more skewed distribution. However, due to the multifaceted nature of imbalance issues in graphs, the specific form of the imbalance ratio can vary. Next, we will introduce the concept of imbalanced learning on graphs as well as the imbalance ratio for graphs.

\begin{table*}[!t]
    \centering
   \small
    \addtolength{\tabcolsep}{-1mm}
    \resizebox{1.0\linewidth}{!}{
    \begin{NiceTabular}{@{}c|c|c|c|p{7cm}@{}} 
    \toprule
     \textbf{Imbalance Types} & \multicolumn{1}{c|}{\textbf{Imbalance Tasks}} & \textbf{Settings} & \textbf{Information Abundance} $s$ & \multicolumn{1}{c}{\textbf{Explanations}}  \\ \midrule\midrule
     \Block{2-1}{\makecell[c]{Node-Level \\ Class \\ Imbalance}} & \makecell[c]{Imbalanced node classification \\ Node-level anomaly detection} & A set of (or two) \emph{node classes} $\bD=\bigcup_{1\leq i\leq K}\bC_i$ & $|\bC_i|$ (\# labeled nodes in each class $\bC_i$) & \Block{1-1}{Labeled nodes are unevenly distributed across classes.} \\ \cmidrule{2-5} 
     & \makecell[c]{Few-shot node classification \\ Zero-shot node classification} & \makecell[c]{A set of base \emph{node classes} $\bD_{b}=\bigcup_{1\leq i\leq K_1}\bC_i$, \\ and novel \emph{node classes} $\bD_{n}=\bigcup_{K_1< i\leq K_2}\bC_i$} & $|\bC_i|$ (\# labeled nodes in each class $\bC_i$) & \Block{1-1}{Base classes have abundant labeled nodes, while novel classes have few/no labeled nodes.} \\ \midrule
    \Block{2-1}{\makecell[c]{Edge-Level \\ Class \\ Imbalance}} & Few-shot link prediction & \makecell[c]{A set of base \emph{graphs} $\bD_b=\{G_i\}^{K_1}_{i=1}$ and novel \\ \emph{graphs} $\bD_n=\{G_i\}^{K_2}_{i=K_1+1}$, where $G_i=\{\bV_i, \bE_i\}$} & $|\bE_i|$ (\# edges in each graph $G_i$) & \Block{1-1}{Base graphs have abundant edges, while novel graphs have limited edges.} \\ \cmidrule{2-5}
    & Edge-level anomaly detection & Two \emph{edge classes} $\bD=\bC_1\cup \bC_2$ & $|\bC_i|$ (\# labeled edges in each class $\bC_i$) & \Block{1-1}{Labeled edges are unevenly distributed across classes.} \\ \midrule
    \Block{2-1}{\makecell[c]{Graph-Level \\ Class \\ Imbalance}} & \makecell[c]{Imbalanced graph classification \\ Graph-level anomaly detection} & A set of (or two) \emph{graph classes} $\bD=\bigcup_{1\leq i\leq K}\bC_i$ & $|\bC_i|$ (\# labeled graphs in each class $\bC_i$) & \Block{1-1}{Labeled graphs are unevenly distributed across classes.} \\ \cmidrule{2-5}
    & Few-shot graph classification & \makecell[c]{A set of base \emph{graph classes} $\bD_{b}=\bigcup_{1\leq i\leq K_1}\bC_i$, \\ and novel \emph{node classes} $\bD_{n}=\bigcup_{K_1< i\leq K_2}\bC_i$} & $|\bC_i|$ (\# labeled graphs in each class $\bC_i$) & \Block{1-1}{Base classes have abundant labeled graphs, while novel classes have few labeled graphs.} \\ \midrule\midrule
    \Block{3-1}{\makecell[c]{Node-Level \\ Structure \\ Imbalance}} & Imbalanced node degrees & \makecell[c]{A set of \emph{node groups} $\bD=\bigcup_{1\leq i\leq K}\bG_i$, where \\ $\bG_i=\{v_j: d_{j}=i\}$ ($d_{j}$ is the degree of node $v_j$)} & $d_j$ (the degree of each node $v_j$) & \Block{1-1}{Head nodes have high degrees, while tail/cold-start nodes have few/no degrees.} \\ \cmidrule{2-5}
    & Node topology imbalance & A set of \emph{node classes} $\bD=\bigcup_{1\leq i\leq K}\bC_i$ & \makecell[c]{The consistency between true class boundaries \\ and influence boundaries of labeled nodes} & \Block{1-1}{Classes with more consistent boundaries tend to propagate label information more effectively.} \\ \cmidrule{2-5}
    & Long-tail entity embedding & \makecell[c]{A set of \emph{entity groups} $\bD=\bigcup_{1\leq i\leq K}\bG_i$, where \\ $\bG_i=\{e_j: d_{j}=i\}$ ($d_{j}$ is the \# triplets of entity $e_j$)} & $d_j$ (\# triplets of each entity $e_j$) & \Block{1-1}{Head entities have more triplets, while tail/cold-start entities have few/no triplets.} \\ \midrule
    \Block{1-1}{\makecell[c]{Edge-Level \\ Structure \\ Imbalance}} & \makecell[c]{Few-shot relation classification \\ Zero-shot relation classification \\ Few-shot reasoning on KGs} & \makecell[c]{A set of base \emph{relations} $\bD_{b}=\bigcup_{1\leq i\leq K_1}\bR_i$, \\ and novel \emph{relations} $\bD_{n}=\bigcup_{K_1< i\leq K_2}\bR_i$} & $|\bR_i|$ (\# labeled triplets of each relation $\bR_i$) & \Block{1-1}{Base relations have abundant labeled triplets, while novel relations have few/no labeled triplets.} \\ \midrule
    \Block{2-1}{\makecell[c]{Graph-Level \\ Structure \\ Imbalance}} & Imbalanced graph sizes & \makecell[c]{A set of \emph{graph groups} $\bD=\bigcup_{1\leq i\leq K}\bG_i$,  where \\ $\bG_i=\{G_j: |\bV_j|=i\}$ ($|\bV_j|$ is the size of graph $G_j$)} & $|\bV_j|$ (the size of each graph $G_j$) & \Block{1-1}{Head graph have large sizes, while tail graphs have small sizes.} \\ \cmidrule{2-5}
    & Imbalanced topology groups & A set of \emph{topology motifs} $\bD=\bigcup_{1\leq i\leq K}\bM_i$ & $|\bM_i|$ (\# instances of each motif $\bM_i$ in one class) & \Block{1-1}{Motifs with more instances have stronger associations with the class than the less frequent motifs.} \\ \bottomrule
     \end{NiceTabular}}
     \caption{Category of existing graph imbalance issues.} \label{table.graph-imba-issues}
\end{table*}

\stitle{Imbalanced learning on graphs.}
Imbalanced learning on graphs is complex due to its distinct graph structures. In addition to class imbalance, the graph structure itself is a significant source of imbalance, which can cause varying performance across groups \wrt their structures.

\begin{definition}[Imbalanced Learning on Graphs]\label{def.imba-learn-graph}
In addition to the number of labeled instances, imbalance in graph data can stem from disparities in structural abundance across groups, leading to a more complex imbalance pattern. For a given graph dataset comprising a set of elements (\ie, nodes, edges, or (sub)graphs) represented as $\bG=\{x_i\}^N_{i=1}$, these elements can be further grouped into $K$ subsets, \ie, $\bG=\bigcup_{1\leq i\leq K}\bG_i$, according to specific criteria based on classes or structures, where $1<K\leq N$. It is important to note that these groups differ in terms of information abundance, which results in unequal performance among them when used as input for a learning model.
\end{definition}

Deviating from the initially defined imbalance ratio for class imbalance tasks, the multifaceted nature of imbalance issues inherent in graphs calls for a broader definition of the imbalance ratio. Formally, given $K$ subsets of a set of elements $\bG=\{x_i\}^N_{i=1}$, \ie, $\bG=\bigcup_{1\leq i\leq K}\bG_i$, let $s_{\bG_i}$ symbolize the abundance of information for group $\bG_i$ based on particular criteria related to classes or structures (as depicted in Table~\ref{table.graph-imba-issues}). In this context, groups are sorted by descending order of information abundance (\ie, if $i<j$, then $s_{\bG_i}\ge s_{\bG_j}$). Thus, the \emph{imbalance ratio on graphs} can be defined as $s_{\bG_1}/s_{\bG_K}$. In essence, this imbalance ratio on graphs quantifies the skewness in distribution across groups concerning their information abundance—a larger ratio denotes a more skewed distribution. Crucially, this generalized definition of the imbalance ratio provides a more flexible tool for evaluating the wide array of imbalance issues within graphs.

To better grasp graph imbalance issues, we have compiled them in Table~\ref{table.graph-imba-issues} to demonstrate their commonalities and disparities.
For each type of imbalance, we enumerate the associated imbalance tasks, all of which are covered in this survey. Based on these tasks, we then outline the main settings pertinent to each task. Furthermore, we introduce the concept of information abundance $s$, as discussed earlier, to pinpoint the imbalance resources that contribute to the imbalance issues. Lastly, for clarity, we furnish detailed explanations for each task.

\section{Overview of Taxonomies} \label{sec.overview}

We categorize the literature from the perspectives of problems and techniques to provide a comprehensive overview.

\stitle{Taxonomy of Problems.}
In terms of the taxonomy of \emph{problems}, our summarization begins by investigating the research tasks of \name. 
Imbalance issues on graphs are multifaceted, stemming from the multi-faceted nature of graph data. Specifically, graph imbalance can arise from two key sources: \emph{classes} and \emph{structures}, which usually serve as the key inputs to learning models. 
On one hand, for classes in graphs, labeled data may be distributed unevenly across classes, resulting in imbalanced outcomes in tasks such as node classification \cite{shi2020multi,zhao2021graphsmote}.
On the other hand, the unique graph structures can introduce another source of imbalance in graph learning models. For instance, the neighborhood abundance of nodes \cite{liu2021tail,zheng2022cold}, can result in discrepancies in the exposed information, which may further skew the output of the learning model accordingly.

Therefore, we first categorize graph imbalance issues based on their types, \ie, \emph{class} and \emph{structure imbalance}, both arising from imbalanced input. Secondly, considering the specific graph structures, we further categorize the imbalance issues into finer groups based on the graph-related elements, \ie, \emph{node-}, \emph{edge-}, and \emph{graph-level imbalance}, to gain a more detailed perspective on these issues. 
Finally, \name\ includes a variety of research tasks, which we use as the finest categories for illustration, such as imbalanced node classification \cite{liu2021pick,shi2020multi,qu2021imgagn}, few-shot node classification \cite{zhou2019meta,yao2020graph,ding2020graph}, \etc\
Fig.~\ref{fig.taxonomy-problems} illustrates the taxonomy of problems.

\stitle{Taxonomy of Techniques.}
For the taxonomy of \emph{techniques}, we categorize the literature based on imbalance forms (\ie, what kind of imbalance) and the corresponding strategies to mitigate the imbalance (\ie, how this imbalance is addressed), as summarized in Fig.~\ref{fig.taxonomy-techniques}.
This is because specific forms of imbalance may require particular techniques to effectively address the associated issues, and the techniques employed may vary across different imbalance forms.
Initially, we separate data into high- and low-resource segments (as shown in Fig.~\ref{fig.intro-imba-problem}(b)) based on their information abundance, which results in two primary task branches contingent on the prediction target: \emph{improving the low-resource part}, and \emph{balancing the high/low-resource parts}. 
The former seeks to improve the performance of the low-resource part without considering the other part; while the latter strives for a balance in performance across both parts, given that predictions will be drawn from both.
Subsequently, we summarize the major techniques employed to address the imbalance issues in each branch, and further provide detailed illustrations of techniques under each main category.

\stitle{Organization.}
In the following three sections, we introduce the two taxonomies of \name. Specifically, Sections~\ref{sec.problem-class} and \ref{sec.problem-structure} address the problems of \name\ from the standpoints of class imbalance and structure imbalance, respectively. In Section~\ref{sec.techniques}, we classify the relevant literature based on the techniques used to address imbalance issues and offer guidance to help readers choose appropriate solutions. Subsequently, we discuss other studies related to imbalanced learning on graphs in Section~\ref{sec.other-literature}.
In Section~\ref{sec.future-directions}, we explore future research directions from the perspective of both problems and techniques, and Section~\ref{sec.conclusions} concludes the survey.

\section{\name\ Problems: Class Imbalance} \label{sec.problem-class}

\begin{figure}[t]
\centering
\includegraphics[width=0.99\linewidth]{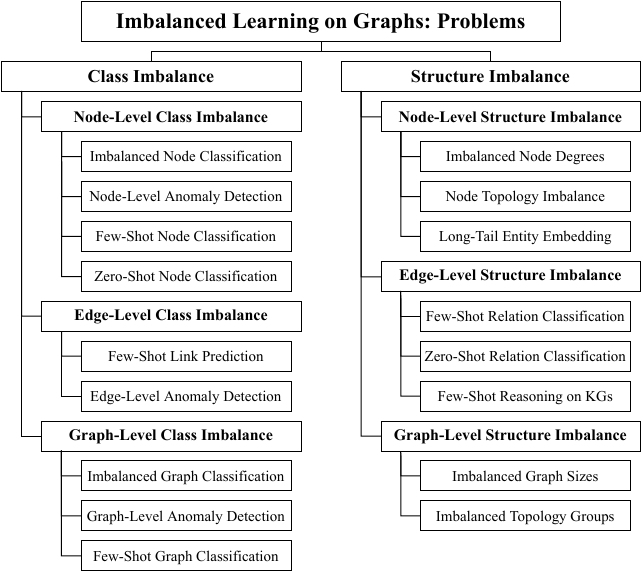}
\caption{Taxonomy of Problems.}
\label{fig.taxonomy-problems}
\end{figure}

In the vast landscape of \name, this section offers a succinct overview of class imbalance issues at the node (Section~\ref{sec.node-class-imba}), edge (Section~\ref{sec.edge-class-imba}), and graph levels (Section~\ref{sec.graph-class-imba}), as illustrated in the left segment of Fig.~\ref{fig.taxonomy-problems}. 
Moreover, structure imbalance will be discussed in Section~\ref{sec.problem-structure}.

\subsection{Node-Level Class Imbalance} \label{sec.node-class-imba}

Node-level class imbalance refers to the disproportionate distribution of labeled nodes across classes. The model often learns more from classes with numerous labeled instances, at the risk of overlooking those with fewer. This challenge surfaces as four principal node-level tasks, which we will detail in the following sections.

\subsubsection{Imbalanced Node Classification} \label{sec.imba-node-class}

Imbalanced node classification is a common challenge in real-world applications where labeled nodes are unevenly distributed across classes (as depicted in Table~\ref{table.graph-imba-issues}), such as in fake account detection or abusive review detection \cite{he2009learning}. This imbalance usually leads to a performance bias, as majority (high-resource) classes tend to outperform minority (low-resource) classes due to their larger number of labeled nodes \cite{shi2020multi,zhao2021graphsmote}. This imbalance has incited a growing number of studies \cite{shi2020multi,zhao2021graphsmote,qu2021imgagn}, trying to enhance the performance of minority groups. For clarity, we categorize these studies into two perspectives: algorithm-level and data-level solutions.

For \emph{algorithm-level} strategies, early work like DRGCN \cite{shi2020multi} implements a GNN-based approach, featuring a generative adversarial network (GAN) \cite{goodfellow2014generative} for synthetic node generation to level the class distribution, paired with a KL-divergence constraint to synchronize the representation distribution of unlabeled nodes with labeled ones.
Differently, DPGNN \cite{wang2021distance} employs a class prototype-driven training approach to balance training loss across classes, with the help of distance metric learning to accurately capture the relative positions of nodes concerning class prototypes.
TAM \cite{song2022tam} addresses the class imbalance issue by accounting for the reduced homogeneity among minority nodes. Specifically, it innovatively introduces connectivity- and distribution-aware margins to guide the model, emphasizing class-wise connectivity and neighbor-label distribution.
LTE4G \cite{yun2022lte4g} considers the imbalance in both node classes and degrees. It partitions nodes into balanced subsets assigned to expert GNNs, then employs knowledge distillation to train class-specific students, to enhance classification performance.

The \emph{data-level} approaches are typically designed to balance data distribution through tail class oversampling \cite{chawla2002smote}, head class undersampling \cite{drummond2003c4}, or instance-based class reweighting \cite{cui2019class}.
On graphs, GraphSMOTE \cite{zhao2021graphsmote} and ImGAGN \cite{qu2021imgagn} both revolve around the generation of synthetic nodes to balance classes. GraphSMOTE achieves this through a technique inspired by SMOTE \cite{chawla2002smote}, generating new node representations by averaging two sampled minority class nodes. ImGAGN, in contrast, applies GAN to generate synthetic nodes.
GraphMixup \cite{wu2022graphmixup}, GraphENS \cite{park2022graphens}, and GraphSANN \cite{liu2023imbalanced} offer a unique approach to synthetic node generation, drawing inspiration from Mixup \cite{zhang2018mixup}. They create new nodes by merging semantic features with contextual edges, promoting balanced model training. Notably, GraphENS places greater emphasis on minority node neighbors, acknowledging their informational bias and employing neighbor sampling along with saliency-based node mixing to mitigate it.
ALLIE \cite{cui2022allie} deviates from synthetic node generation, employing active learning instead. It involves sampling instances from both majority and minority classes, with an imbalance-aware reward function implemented via reinforcement learning \cite{kaelbling1996reinforcement} aiding the sampling process.

\begin{table}[!t]
    \centering
   \small
    \addtolength{\tabcolsep}{-1mm}
    \resizebox{1.0\linewidth}{!}{
    \begin{NiceTabular}{@{}c|c||r@{}} 
    \toprule
     \multicolumn{2}{c||}{\textbf{Techniques}} & \multicolumn{1}{c}{\textbf{Literature}}  \\ \midrule\midrule
     \multirow{2}{*}{{\centering Algo-level}} & Constraints & DRGCN \cite{shi2020multi}, DPGNN \cite{wang2021distance}, TAM \cite{song2022tam} \\ 
    & Knowledge distillation & LTE4G \cite{yun2022lte4g} \\ \midrule
     \multirow{5}{*}{{\centering Data-level}} & GAN &  DRGCN \cite{shi2020multi}, ImGAGN \cite{qu2021imgagn}   \\ 
     & SMOTE &  GraphSMOTE \cite{zhao2021graphsmote}  \\
     & Mixup &  GraphMixup \cite{wu2022graphmixup}, GraphSANN \cite{liu2023imbalanced}, GraphENS \cite{park2022graphens} \\ 
     & Resampling & LTE4G \cite{yun2022lte4g}, ALLIE \cite{cui2022allie} \\
      & Reweighting & TAM \cite{song2022tam} \\
    \bottomrule
     \end{NiceTabular}}
     \caption{Summary of imbalanced node classification.} \label{table.imba-node-class}
\end{table}

\stitle{Summary.}
The core challenge in imbalanced node classification lies in \emph{achieving balanced information distribution across classes for uniform model training}. 
Algorithm-level methods addressing imbalance often leverage strategies such as additional constraints \cite{shi2020multi}, prototype-driven training \cite{wang2021distance}, ensuring homogeneity among minority nodes \cite{song2022tam}, or employing knowledge distillation \cite{yun2022lte4g}.
As summarized in Table~\ref{table.imba-node-class}, a significant portion of research in this area tackles this issue from the data-level. This commonly involves creating synthetic nodes for underrepresented classes, or opting for reweighting or resampling nodes from both the majority and minority classes. 
For a more in-depth understanding of these techniques, please refer to Section~\ref{sec.balan-both-parts}.
Note that, there is no significant gap between algorithm- and data-level approaches, as some methods, like DRGCN and TAM, can incorporate elements from both.
Despite extensive research, this area still calls for further exploration. Innovative techniques, such as using diffusion models \cite{rombach2022high} to generate synthetic nodes or employing transfer learning to transfer knowledge from majority to minority classes for their enhancement, could be promising avenues of investigation.

A particular type of imbalanced node classification, node-level anomaly detection, has sparked increasing interest recently. We will separately provide a comprehensive review of the associated literature in Section~\ref{sec.node-level-anomaly-detection}.

\eat{
Most studies try to generate fake nodes for the minority classes, by such as GAN \cite{shi2020multi,qu2021imgagn}, SMOTE \cite{zhao2021graphsmote}, or Mixup \cite{park2022graphens,wu2021graphmixup}; some attempt to reweight the nodes \cite{song2022tam}, and the others try to resample the instances for the classes \cite{cui2022allie}. Overall, they follow the main directions to cope with class imbalance and further incorporate graph attributes into their algorithms.
Note that, as a specific form of imbalanced node classification, node-level anomaly detection has drawn the increasing attention of researchers recently. Due to its particularity, we separately summarize the literature for node-level anomaly detection in the following part.
}

\subsubsection{Node-Level Anomaly Detection} \label{sec.node-level-anomaly-detection}  

Node-level anomaly detection, as a specific manifestation of imbalanced node classification, generally centers on binary classification between normal and anomalous nodes. This task plays a vital role in numerous applications, such as outlier detection \cite{qiu2022raising,ma2022deep}, fraudulent user detection \cite{zhao2021action,zhang2021fraudre,qian2021distilling,mathew2021defraudnet}, and social spammer detection \cite{wang2019nodes,miz2019anomaly,lin2019fraud}. Considering the diversity of graphs (\eg, homogeneous, heterogeneous, or dynamic graphs), a range of techniques have been developed to address this issue, such as traditional graph algorithms (GA), graph embedding approaches (GE), and graph neural network approaches (GNN). They are categorized in Table~\ref{table.anomaly-detection}, which also includes edge- and graph-level anomaly detection approaches. While we provide an overview of recent literature in this domain, for an in-depth analysis, we refer readers to the comprehensive survey \cite{ma2021comprehensive}.

\begin{table}[!t]
\centering
\small
\addtolength{\tabcolsep}{0pt}
 \resizebox{1\linewidth}{!}{
\begin{NiceTabular}{@{}c|c|c|r@{}}
\toprule
 \textbf{Graph objects} & \textbf{Graph types} & \textbf{Base models} & \textbf{Literature} \\ \midrule\midrule
         \multirow{11}{*}{Node-level} 
         & \multirow{4}{*}{Homogeneous} & GA & \cite{hu2016embedding,li2017radar,wang2017gang,bojchevski2018bayesian,peng2018anomalous,yang2019mining,ding2019interactive,gutierrez2020multi,zhu2020mixedad,aggarwal2021signature,nguyen2023detecting} \\ 
         &  & GE & \cite{yuan2017spectrum,liang2018semi,zhou2018sparc,bandyopadhyay2019outlier,bandyopadhyay2020integrating,ahmed2021graph} \\ 
         &  & GNN & \cite{li2019specae,ding2019deep,zhao2020error,peng2020deep,chen2020generative,cheng2020graph,ding2021cross,deng2021graph,desai2021graph,jin2021anemone,ding2021few,zhou2021subtractive,ding2021inductive,liu2021anomaly,mathew2021defraudnet,zhao2021synergistic,chen2022deep,qi2022mad,chen2022graphad,zhang2022reconstruction,luo2022comga,goodge2022lunar,zhou2022unseen,liu2022dagad,huang2022hop,zhuang2023subgraph,gao2023addressing,huang2023unsupervised} \\ \cmidrule{2-4}

         & \multirow{3}{*}{Heterogeneous} & GA & \cite{ying2016pfraudetector,liu2017accelerated,hooi2017graph,zhang2017hidden,cao2017hitfraud,ranjbar2018qanet,ren2021ensemfdet} \\ 
         &  & GE & \cite{fan2018gotcha,yoon2021graph} \\ 
         &  & GNN & \cite{dou2020enhancing,zhu2020heterogeneous,liu2020alleviating,zhang2020gcn,qian2021distilling,zhang2021fraudre,hu2021turbo,xu2021towards,liu2021pick,wang2021modeling,wang2021decoupling,liu2021intention,dong2022bi,huang2022auc,li2022devil,qin2022explainable,tang2022rethinking,shi2022h2,gao2023addressing,gao2023alleviating} \\ \cmidrule{2-4}
         
         & \multirow{4}{*}{Dynamic} & GA & \cite{rashidi2016node,liu2017holoscope,fond2018designing,liu2018contrast,sharma2018nhad,lin2019fraud,yoon2019fast,bhatia2020midas,guo2022subset} \\ 
         &  & GE & \cite{teng2017anomaly,yu2018netwalk,teng2018deep,chen2020anomaly,ofori2021topological} \\ 
         &  & GNN & \cite{wang2021bipartite,zhao2021action,deng2022graph,lu2022bright,cui2022meta,tian2023sad} \\ 
         \midrule \midrule
         
         \multirow{6}{*}{Edge-level} & Homogeneous & GNN & \cite{duan2020aane,liu2022deep} \\ \cmidrule{2-4}
         & Heterogeneous & GNN & \cite{li2021live} \\ \cmidrule{2-4}
         & \multirow{3}{*}{Dynamic} & GA & \cite{wang2019nodes,chang2021f} \\ 
         &  & GE & \cite{miz2019anomaly} \\ 
         &  & GNN & \cite{zheng2019addgraph,cai2021structural} \\ 
         \midrule \midrule
         
         \multirow{5}{*}{Graph-level} 
         & \multirow{2}{*}{Homogeneous} & GA & \cite{wu2017query,perozzi2018discovering,wu2019uncovering,sun2020anomaly,wang2022calibrated,chen2022antibenford} \\ 
         &  & GNN & \cite{bhatia2021graphanogan,qiu2022raising,ma2022deep,zhang2022unsupervised} \\ 
         \cmidrule{2-4}
        
         & Heterogeneous & GA & \cite{hou2023efficient} \\ \cmidrule{2-4}
         & \multirow{2}{*}{Dynamic} & GA & \cite{manzoor2016fast,zambon2018concept} \\
         &  & GNN & \cite{zhao2020multivariate} 
         
 \\\bottomrule
\end{NiceTabular}}
\caption{Summary of anomaly detection on graphs. } \label{table.anomaly-detection}
\end{table}

\stitle{Homogeneous graphs.} 
Node-level anomaly detection on homogeneous graphs is a focal point of interest within this task. Traditional graph algorithms (GA) address anomaly detection by capturing structural anomalies through various handcrafted metrics such as applying residual analysis on attributes \cite{li2017radar,peng2018anomalous}, or utilizing human feedback in an interactive manner \cite{ding2019interactive}. Following this, graph embedding (GE) based methods were developed, aiming to separately embed anomalous and normal nodes to capture their specific patterns \cite{liang2018semi,bandyopadhyay2019outlier}, or to generate representations that can form clear boundaries between classes \cite{bandyopadhyay2020integrating}, among others.

Recently, GNNs are widely used for anomaly detection on graphs. These methods are usually coupled with different learning objectives to obtain distinguishable node embeddings. We outline some popular techniques below. 
Contrastive learning methods typically leverage the similarity between target nodes and their context during model pre-training. For graph data containing suspicious nodes---nodes that often attempt to camouflage themselves---identifying their dissimilarity to the context (\ie, heterophily) proves beneficial for their detection \cite{liu2021anomaly,jin2021anemone}. Generative models like GANs are also exploited to create synthetic nodes during training, so the model can learn the distinct distributions of normal nodes as well as the out-of-distribution anomalous nodes \cite{chen2020generative,ding2021inductive}. In addition, autoencoders are also employed to reconstruct graph structures or node features, where hard-to-reconstruct samples are usually identified as anomalies \cite{chen2020generative,desai2021graph}.

\stitle{Heterogeneous graphs.} 
Anomaly detection is extensively studied in the context of heterogeneous graphs as well. In early stages, traditional graph algorithms (GA) usually define anomalous structural patterns on heterogeneous graphs to identify anomalies, such as detecting dense subgraphs \cite{hooi2017graph,zhang2017hidden}, or finding suspicious node attributes in clusters \cite{liu2017accelerated}. Later in the literature, more GE-based methods are proposed for heterogeneous graphs, such as characterizing different types of malware using meta-path based random walks \cite{fan2018gotcha}, or identifying fraudsters based on their propensity to form unidirectional links with normal users \cite{yoon2021graph}.

In addition, many recent studies leverage heterogeneous GNNs to concurrently address graph heterogeneity and imbalance challenges, employing a variety of effective techniques. For instance, some studies \cite{dou2020enhancing,huang2022auc} resort to reinforcement learning to select the most informative neighbors, and jointly consider multiple relations to learn distinguishable node embeddings. Similarly, some studies \cite{liu2020alleviating,liu2021pick} utilize neighborhood sampling techniques to tackle class imbalance and aggregate multiple relations to differentiate normal and anomalous neighboring nodes. Additionally, spectral-based methods \cite{wang2021decoupling,tang2022rethinking} segregate aggregations of normal and anomalous nodes using distinct pass filters and manage multiple relations individually.

\stitle{Dynamic graphs.} 
Node-level anomaly detection is also widely explored within the realm of dynamic graphs.
Early GA approaches address anomaly detection on dynamic graphs typically by identifying changes or unusual patterns within the graph structure, such as detecting temporal bursts \cite{liu2017holoscope} or tracking sudden changes on anomaly scores \cite{yoon2019fast}. 
Additionally, GE-based methods are proposed to detect graph anomalies by learning distinguishable node embeddings, such as applying support vector data description (SVDD) to differentiate time-sensitive anomalous patterns \cite{teng2017anomaly}, using autoencoder to learn dynamic node embeddings and detect anomalous nodes by their distance to cluster centers \cite{yu2018netwalk}, or combining autoencoders with hypersphere learning to isolate the spatiotemporal anomalies \cite{teng2018deep}.

Several GNN-based methods are also proposed to address anomaly detection on dynamic graphs \cite{wang2021bipartite,zhao2021action,deng2022graph,lu2022bright,cui2022meta,tian2023sad}. For example, STGAN \cite{deng2022graph} involves a spatiotemporal sequence generator and a discriminator that determines whether an input sequence is real or not. During inference, the discriminator reports samples that have low scores as anomalies due to their out-of-distribution representations. 
Additionally, AddGraph \cite{zheng2019addgraph} leverages an extended temporal graph convolutional network to discern both long-term and short-term patterns in dynamic graphs, using a margin loss to learn distinctive representations for anomalous and normal nodes.

\stitle{Summary.}
Node-level anomaly detection on graphs is inherently challenging due to the intricate nature of graph structures and the infrequency of anomalies. Prevalent techniques in this arena largely focus on establishing distinct boundaries between normal and anomalous nodes. This can be achieved through feature engineering (GA), discriminative node representations (GE), or neighborhood aggregation for structure encoding (GNN). For a more comprehensive insight into state-of-the-art comparisons, readers may refer to benchmarks such as \cite{liu2022bond} or visit leaderboards\footnote{\url{https://dgraph.xinye.com/leaderboards/dgraphfin}.}.
To further propel progress in this domain, embracing innovative methodologies is essential. For example, sophisticated generative learning models, like diffusion \cite{rombach2022high}, can be harnessed to bolster synthetic instance generation, thereby enhancing the discernibility of anomalous nodes. Furthermore, innovative foundational models \cite{bommasani2021opportunities} hold promise for enriching the comprehension of graph structures, and thus can potentially facilitate more effective anomaly detection.

\subsubsection{Few-Shot Node Classification}
In various real-world settings, \emph{novel classes} are often encountered, marked by the availability of only a few labeled instances or \emph{few-shot} instances. Such scarcity imposes significant hurdles for model training. To counter this, a set of \emph{base classes} abundant in labeled data is typically harnessed to assist the learning process \cite{wang2020generalizing}, as summarized in Table~\ref{table.graph-imba-issues}. Thus, the primary objective of \emph{Few-Shot Node Classification} (FSNC) on graphs is to yield robust and superior performance on these few-shot novel classes by efficiently exploiting the information derived from the base classes, and this task has recently garnered considerable attention \cite{zhou2019meta,wang2020graph,liu2021relative,zhou2022task}. 
This task finds applications in several contexts, such as novel intrusion detection on traffic networks, predicting new types of goods in e-commerce, and anticipating newly discovered chemical properties in protein networks, \etc\
In this section, we will delve into FSNC across various settings.

\begin{table*}[!t] 
    \centering
    \small
    \addtolength{\tabcolsep}{-1.5pt}
   \resizebox{1\linewidth}{!}{%
    \begin{NiceTabular}{@{}l||c|c|c|c|c|c@{}}
    \toprule
  \Block[c]{2-1}{\textbf{Tasks}} & \multicolumn{3}{c|}{\textbf{Meta-learning techniques}} & \multicolumn{3}{c}{\textbf{Other techniques}}\\ \cmidrule{2-7}
  & MAML & Prototypical network & Others & Label generation & Contrastive Learning & Prompting \\\midrule\midrule
    Generic FSNC & \cite{zhou2019meta,wang2020graph,liu2021relative,zhou2022task}  & \cite{yao2020graph,ding2020graph,huang2020graph,wang2022task,liu2022few,zhang2022mul} & - & \cite{ijcai2022p500} & \cite{tan2022supervised,tan2022transductive} & \cite{liu2023graphprompt} \\ \midrule
    Generalized FSNC & - & \cite{tan2022graph,lu2022geometer} & \cite{xu2022generalized} & - & - & - \\ \midrule
    Multi-label FSNC & - & \cite{lan2020node} & - & - & - & -  \\ \midrule
    FSNC with extremely weak supervision & - & - & - & \cite{wangfewshotnode2023} & - & -    \\ \midrule
    FSNC on HINs & \cite{zhang2022few,zhang2022hg} & - & - & - & - & -  \\ 
    \bottomrule
     \end{NiceTabular}}
     \caption{Summary of few-shot node classification.}
      \label{table.few-shot-node-class}%
\end{table*}

\stitle{Generic FSNC.}
Meta-learning \cite{hospedales2021meta}, with a primary focus on MAML \cite{finn2017model} and prototypical networks \cite{snell2017prototypical}, is a common approach in many studies addressing FSNC.

\ititle{MAML-based approaches.}
Meta-GNN \cite{zhou2019meta} set a precedent in the field by utilizing meta-learning to transfer meta-knowledge from data-rich base classes to novel classes using a MAML-based adaptation \cite{finn2017model}.
Building on this, AMM-GNN \cite{wang2020graph} introduces attribute-level adaptation using FiLM \cite{perez2018film} to address potential attribute disparities across meta-tasks, complementing the MAML adaptation.
TRGM \cite{zhou2022task} adds another dimension by capturing inter-task relations via contrastive learning in a constructed task graph, thereby enhancing few-shot learning.
In a distinct approach, RALE \cite{liu2021relative}, while rooted in MAML, emphasizes both relative and absolute node positions for node placement, aiming to bolster representations of nodes in novel classes.

\ititle{Prototypical network-based approaches.}
GFL \cite{yao2020graph} utilizes prototypical network to enable cross-graph FSNC, and further introduces a hierarchical graph representation gate to regulate global transferable knowledge.
Moreover, GPN \cite{ding2020graph} enriches this paradigm by incorporating a node valuator to re-weight labeled nodes according to their informativeness in prototype calculation.
In addition, MuL-GRN \cite{zhang2022mul}, akin to RALE \cite{liu2021relative}, calculates relations at node, global subgraph, and local subgraph levels between query and support nodes to facilitate the classification of novel class nodes.

Some approaches merge the benefits of prototypical networks with adaptation-based methods like MAML or FiLM.
ST-GFSL \cite{lu2022spatio} and G-Meta \cite{huang2020graph}, building on the prototypical network, employ MAML for optimization to tackle FSNC, with the former addressing traffic flow prediction on spatio-temporal graphs and the latter emphasizing local subgraph utilization for target node representation.
TENT \cite{wang2022task} tackles variance across meta-tasks via node-, class-, and task-level adaptations.
Moreover, Meta-GPS \cite{liu2022few} refines aggregation and combination functions in GNNs to consider both graph homophily and heterophily, employing FiLM \cite{perez2018film} to modulate model parameters per meta-task.

\ititle{Other techniques.}
While most existing studies employ meta-learning for FSNC, alternative techniques have also been proposed.
For instance, IA-FSNC \cite{ijcai2022p500} initializes novel classes with the first layer of a pre-trained GNN model, and employs a semi-supervised method for synthetic label generation to realize information augmentation for the support sets.
In addition, Tan \etal\ \cite{tan2022supervised,tan2022transductive} employ contrastive learning, by constructing a subgraph around a target node based on connectivity and generating three types of contrast pairs (node-node, node-subgraph, and subgraph-subgraph) for model training.
Another work GraphPrompt \cite{liu2023graphprompt} employs GNN prompting techniques, unifying pre-training and downstream tasks for effective knowledge transfer, achieving impressive results even with limited supervision.

\stitle{Generalized FSNC.}
In a broader FSNC setting, both base and novel class nodes appear in the test set rather than only novel classes. This complicates matters by introducing challenges like asymmetric classification and inconsistent preference between base and novel classes \cite{xu2022generalized}.
Stager \cite{xu2022generalized} addresses this by using prediction uncertainty to measure class shot levels and decompose prediction probability, tackling asymmetric classification. It also employs a meta-learner to weigh various receptive fields, addressing the inconsistent preference challenge.

Other works \cite{tan2022graph,lu2022geometer} focus on class-incremental FSNC, where novel classes continually emerge. 
Both HAG-Meta \cite{tan2022graph} and Geometer \cite{lu2022geometer} address this challenge by establishing an attention-based framework on the foundation of the prototypical network, allowing for the weighted calculation of prototypes from labeled instances.
HAG-Meta extends its approach beyond GNN pre-training on base classes, using an attention mechanism to adjust the significance of meta-tasks. On the other hand, Geometer applies geometric loss functions to maintain intra-class proximity, inter-class uniformity, and inter-class separability, effectively tackling the challenge of class incrementality.

\stitle{Multi-label FSNC.}
While the majority of prior research primarily focuses on multi-class node classification, the task of few-shot multi-label node classification, where each node may be associated with multiple labels instead of just one, continues to pose a considerable challenge. 
To address this issue, MetaTNE \cite{lan2020node} treats it as a label-wise binary classification problem. It identifies discrepancies in node embeddings across different meta-tasks and implemented an embedding transformation. This transformation, contingent on the query node within each task, allows it to obtain task-specific representations beneficial for classification.

\stitle{FSNC with extremely weak supervision.}
In the challenging scenario of FSNC with extremely weak supervision, where only a limited number of labeled nodes are available during meta-training, models typically struggle to extract sufficient prior knowledge for effective knowledge transfer. To overcome this limitation, X-FNC \cite{wangfewshotnode2023} introduces a novel approach to generate pseudo-labels as additional references for the base classes, in order to learn effectively from extremely weak supervision.

\stitle{FSNC on HINs.}
Recent research has delved into FSNC on heterogeneous graphs, necessitating the management of heterogeneity. 
To enhance classification in the target domain, CrossHG-Meta \cite{zhang2022fewheterogeneous}, building on heterogeneous graph neural networks (HGNN) \cite{wang2019heterogeneous,lv2021we}, extracts transferable meta-knowledge from a source domain. It employs dual adaptations at the domain- and task-levels for model optimization, along with a cross-domain contrastive regularization.
Similarly, HG-Meta \cite{zhang2022hg}, another study founded on HGNN, tackles differences across meta-tasks through task-level modulation. It applies task feature scaling to adjust node representations and incorporates an attention mechanism to re-evaluate the importance of each meta-task.

\stitle{Summary.}
The primary challenge in few-shot node classification is to \emph{extract transferable knowledge from base classes to benefit novel classes}. Recent studies in this area, as summarized in Table~\ref{table.few-shot-node-class}, have focused on various aspects of FSNC, including generic, generalized (class incremental), multi-label, extremely weak supervision settings, and on HINs. Methods frequently employed include meta-learning techniques like MAML and prototypical networks, sometimes used simultaneously. Yet, other studies have also ventured into utilizing innovative techniques such as data augmentation, contrastive learning, and prompting methods.
For a more comprehensive understanding of these techniques, we direct readers to Section~\ref{sec.knowledge-transfer}.
Despite considerable research, this complex task continues to necessitate further exploration. As demonstrated in Table~\ref{table.few-shot-node-class}, specific settings like generalized, multi-label, extremely weak supervision, or those concerning HINs, remain largely underexplored. Notably, the adoption of innovative techniques like prompt tuning \cite{liu2023graphprompt}, generative models \cite{rombach2022high}, and others could potentially enhance the representation of novel classes, indicating promising future directions.
For a broader perspective, we suggest consulting the survey of few-shot learning on graphs \cite{zhang2022few} for additional insights.

\subsubsection{Zero-Shot Node Classification}
Zero-Shot Node Classification (ZSNC) necessitates the absence of labeled data for novel classes during model training. Wang \etal\ \cite{wang2018rsdne} build on DeepWalk \cite{perozzi2014deepwalk}, modulating the objective constraints on base classes to learn effective representations for unlabeled novel classes. They relax the intra-class similarity constraint and remove inter-label edges to enhance dissimilarity. Its further extensions \cite{wang2020network,wang2021expanding} utilize GCN as the backbone, adding label-related semantic descriptions. Differently, DGPN \cite{wang2021zero} provides a formalization for ZSNC, introducing class semantic descriptions and corresponding evaluation metrics (based on the proximity of text-based class embeddings and proximity in real data). They refine GNN models by constraining the similarity between hidden embeddings (in each layer) and the class embeddings, to enhance the model training.
Similarly, another work DBiGCN \cite{yue2022dual} also follows the paradigm of utilizing class descriptions for ZSNC.

\stitle{Summary.}
ZSNC remains an underexplored field due to the absence of descriptions for elements like nodes, edges, or graphs, which are crucial in typical zero-shot settings. Therefore, this challenging task warrants further research attention. Currently, the existing strategies involve constraining model training on base classes, while others utilize auxiliary information, such as descriptions, to supplement model training. The metrics established by previous work, such as \cite{wang2021zero}, lay a promising foundation for further exploration in this challenging domain.

\subsection{Edge-Level Class Imbalance} \label{sec.edge-class-imba}

In addition to node-level class imbalance, addressing edge-level class imbalance is also crucial. Two representative tasks of interest are few-shot link prediction and edge-level anomaly detection. These tasks underscore the importance of tackling edge-level class imbalance in various real-world scenarios, such as inductive recommendation \cite{zhu2023few}, anomalous transaction detection on financial networks, \etc\

\subsubsection{Few-Shot Link Prediction}
Recent studies have focused on few-shot link prediction across graphs, particularly in event-based social networks \cite{zhu2023few}. This presents a challenge as the novel graphs in this context have limited edges and pose difficulties for link prediction. 
To address this, EA-GAT \cite{zhu2023few}, an event-aware graph attention network, attempts to effectively encode fine-grained events from existing graphs and few-shot target events. It further utilizes gradient-based episode learning to acquire transferable knowledge and adapt to event-based social networks with sparse connections.

Additionally, other settings such as few-shot link prediction across different sections of a single graph, which might demonstrate varied patterns \cite{liu2021nodewise}, could also be explored as a potential avenue for research.

\subsubsection{Edge-Level Anomaly Detection}  
Edge-level anomaly detection is designed to identify anomalous edges within a network. This endeavor is crucial in areas like transactions and social networks to detect potentially dubious transactions or fake news. Prevailing research primarily focuses on characterizing edge distributions and pinpointing anomalies that exhibit significant deviations from these established distributions.

Existing works on homogeneous graphs tend to use graph auto-encoders to model the distribution of normal edges and use reconstruction errors as indicators of anomalies \cite{duan2020aane}, or enrich node and edge features using auxiliary descriptions \cite{liu2022deep} for better discrimination.
On heterogeneous graphs, research often tries to capture both heterogeneity and anomalous patterns, such as using type-aware edge representations and tailored classifiers to identify fraudulent transactions on live-streaming platforms \cite{li2021live}.

Some works consider detecting abnormal edges in dynamic graphs by keeping track of edge dynamics \cite{wang2019nodes,miz2019anomaly,zheng2019addgraph,chang2021f,cai2021structural}. For instance, Miz \etal\ \cite{miz2019anomaly} use the Hopfield model of memory to combine graph and temporal information and detect temporal spikes in activities as anomalies. 
Another work StrGNN \cite{cai2021structural} extracts subgraph snapshots around an edge and applies a GNN model with gated recurrent units \cite{chung2014empirical} to capture subgraph structural changes as anomaly indicators. 

\stitle{Summary.} 
The main challenge in edge-level anomaly detection lies in the highly imbalanced distribution of normal and abnormal edges. To counter this challenge, previous works usually model edge distributions based on the abundant information on normal edges, using deviations from the normal distributions as anomaly indicators. Despite its significance, this area remains relatively underexplored. There is ample room for further investigation, especially in tasks involving heterogeneous graphs where edge relations are intricate and significant.

\begin{table*}[!t] 
    \centering
    \small
    \addtolength{\tabcolsep}{-1.5pt}
   \resizebox{1\linewidth}{!}{%
    \begin{NiceTabular}{@{}l||c|c|c@{}}
    \toprule 
  \Block[c]{2-1}{\textbf{Tasks}} & \multicolumn{2}{c|}{\textbf{Meta-learning techniques}} & \Block[c]{2-1}{\textbf{Other techniques}}\\ \cmidrule{2-3}
  & MAML & Prototypical network &  \\\midrule\midrule
    Generic FSGC & \cite{ma2020adaptive}  & \cite{wang2022faith,crisostomi2022metric} & Adaptive step controller \cite{ma2020adaptive}, super-class graph \cite{Chauhan2020FEWSHOT}, task correlations \cite{wang2022faith}  \\ \midrule
    Cross-domain FSGC & - & - & Data augmentation \cite{hassani2022cross}    \\ \midrule
    Few-shot temporal graph classification & \cite{fu2022meta} & \cite{fu2022meta} & -  \\ \midrule
    Few-shot molecular property prediction & \cite{wang2021property,guo2021few,borde2022graph,meng2023meta} & - & Meta-task reweighting \cite{guo2021few}, implicit function theorem \cite{chen2023metalearning} \\\bottomrule
     \end{NiceTabular}}
     \caption{Summary of few-shot graph classification.}
      \label{table.few-shot-graph-class}%
\end{table*}

\subsection{Graph-Level Class Imbalance} \label{sec.graph-class-imba}

Addressing graph-level class imbalance forms a vital aspect of \name, surpassing the scope of node- and edge-level imbalance. Three crucial areas warrant attention within this domain: imbalanced graph classification, few-shot graph classification, and graph-level anomaly detection. They highlight the importance of addressing graph-level class imbalance across a variety of real-world scenarios, such as novel drug detection, and protein function prediction.

\subsubsection{Imbalanced Graph Classification}

Imbalanced graph classification parallels the challenges of imbalanced node classification. It commonly arises in real-world scenarios (\eg, imbalanced chemical compound classification) where class distributions of labeled graphs are skewed (as illustrated in Table~\ref{table.graph-imba-issues}), often favoring the majority class with more labeled graphs \cite{wang2022imbalanced}. 

To counter this, $\text{G}^2\text{GNN}$ \cite{wang2022imbalanced} utilizes additional supervision both globally, through neighboring graphs, and locally, via stochastic augmentations. It builds a Graph of Graphs (GoG) by leveraging kernel similarity and implements GoG propagation for information aggregation. Additionally, topological augmentation paired with self-consistency regularization is employed at the local level. These strategies collectively enhance model generalizability, thereby elevating classification performance.

Despite the advance, this crucial task continues to call for more in-depth study due to the limited amount of current research, indicating a potentially promising direction.

\subsubsection{Graph-Level Anomaly Detection}  

Graph-level anomaly detection aims to detect anomalous graphs or subgraphs according to a predefined anomaly measure, which can be seen as a variant of imbalanced graph classification with binary labels. This task is commonly seen in transaction networks where fraudsters conspire to do money laundering.

Much of the emphasis has been placed on \emph{homogeneous graphs}. A segment of research employs traditional graph algorithms (GA) to identify anomalous subgraphs\cite{wu2017query,perozzi2018discovering,wu2019uncovering,sun2020anomaly,wang2022calibrated,chen2022antibenford}. 
For instance, AMEN \cite{perozzi2018discovering} defines a normality measure and discovers communities with an aim to maximize their normality scores, where communities that cannot achieve high normality scores are considered to be fraudulent. 
Chen \etal\ \cite{chen2022antibenford} employ Benford’s law to assign anomaly scores to edges and use the densest sub-graph discovery algorithm to find anomalous sub-graphs.

Another line of works uses GE-based methods to analyze the anomaly degree of graphs \cite{bhatia2021graphanogan,qiu2022raising,ma2022deep,zhang2022unsupervised}. 
For instance, OCGTL \cite{qiu2022raising} concentrates graph embeddings in the training set within a hypersphere where the distance to the hypersphere is used as the anomaly degree.  
AS-GA \cite{zhang2022unsupervised} identifies suspicious sub-graphs based on the reconstruction error of a graph auto-encoder and rates anomaly scores of extracted areas with a graph supermodular neural network.

Some studies consider graph-level anomaly detection on \emph{heterogeneous graphs}.
For instance, ACGPMiner \cite{hou2023efficient} introduces conditional graph patterns to model abnormal patterns in property graphs and follows a generation-and-validation paradigm to mine the defined abnormal patterns and their matches.

In \emph{dynamic graphs}, existing works usually try to detect the deviation of graph structures among different time steps \cite{manzoor2016fast,zambon2018concept,zhao2020multivariate}. 
For instance, Zambon \etal\ \cite{zambon2018concept} propose to embed each graph in the graph stream into a vector and perform change detection on the converted vector stream to flag unusual graph patterns. 
MTAD-GAT \cite{zhao2020multivariate} employs self-attention to capture temporal relations across different time steps for nodes and aggregates node-level anomaly scores to generate graph-level anomaly scores. 

\stitle{Summary.} 
Graph-level anomaly detection is a logical progression from node- and edge-level tasks. Many current approaches determine node- or edge-level anomaly scores and subsequently aggregate these to gauge graph-level anomalies. An alternative strategy involves embedding entire graphs into vectors, effectively transforming graph-level anomaly detection into the well-established realm of vector-based anomaly detection. Presently, the primary focus lies on homogeneous and dynamic graphs. Nevertheless, detecting anomalies at the graph-level on heterogeneous graphs remains a crucial area meriting further exploration.

\subsubsection{Few-Shot Graph Classification}

Few-Shot Graph Classification (FSGC), a crucial facet of imbalanced learning on graphs, aims to categorize target graphs into respective novel classes, commonly applied in real-world scenarios such as novel molecular or protein prediction.
Like few-shot node classification, the objective of FSGC is to distill transferable knowledge from base classes and apply it to novel classes, thereby enhancing the classification performance for the latter.

\stitle{Generic FSGC.}
To address the challenge of FSGC, several noteworthy studies offer distinct solutions.
AS-MAML \cite{ma2020adaptive} employs MAML \cite{finn2017model} to extract meta-knowledge and uses a reinforcement learning-based adaptive controller to adjust the step size \cite{williams1992simple}. 
In contrast, Chauhan \etal\ \cite{Chauhan2020FEWSHOT} utilize distance metrics to cluster closely related graph classes into super-classes, forming a super-graph. This structure facilitates hierarchical classification at both super-class and class levels, thereby highlighting the importance of inter-class relationships.
Different from the class correlations, FAITH \cite{wang2022faith} focuses on the role of task correlations in knowledge transfer. They build a three-layer hierarchical graph capturing sample-, prototype-, and task-level data to facilitate classification, emphasizing the role of task correlations in model performance.
In addition, Crisostomi \etal\ \cite{crisostomi2022metric} combine a simple metric learning model with advanced graph embedding techniques, demonstrating excellent results. Their strategy of using a task-adaptive mechanism for layer-wise adaptation and Mixup data augmentation showcases an effective blend of traditional and novel techniques for superior performance.

\stitle{Cross-domain FSGC.}
For cross-domain FSGC, Hassani \etal\ \cite{hassani2022cross} devise three graph augmentations, including contextual and two topological augmentations, to enhance representation learning, particularly task-specific information for fast adaptation and task-agnostic information for knowledge transfer. 

\stitle{Few-shot temporal graph classification.}
Fu \etal\ \cite{fu2022meta} explore a novel setting of few-shot temporal graph classification, where each class may have only a few labeled temporal graphs and novel classes might emerge in the future. They present Temp-GFSM \cite{fu2022meta}, a temporal graph metric learning framework. Utilizing attention mechanisms like node-level lifelong attention and both intra- and inter-snapshot attention, they derive temporal graph representations. They are subsequently used in few-shot metric learning, backed by prototypical networks and MAML.

\stitle{Few-shot molecular property prediction.}
In the realm of few-shot molecular property prediction, several studies exploit meta-learning (\eg, MAML) to bridge different tasks. PAR \cite{wang2021property} applies MAML, utilizing detailed molecular representations to enhance prediction performance. Meta-MGNN \cite{guo2021few} also employs MAML, but enhances its approach by applying self-supervised constraints for graph knowledge extraction and an attention mechanism to weight meta-task contributions.
Similarly, Borde \etal\ \cite{borde2022graph} highlight the applicability of Reptile \cite{nichol2018reptile}, a MAML-like model-agnostic algorithm, with GNN models in molecular property regression tasks. 

Additionally, various other techniques are also employed.
ADKF-IFT \cite{chen2023metalearning} integrates meta-learning with deep kernel Gaussian processes \cite{seeger2004gaussian}, proposing adaptive deep kernel fitting for versatile feature representation learning across tasks.
In contrast, to combat overfitting and improve generalizability, MTA \cite{meng2023meta} generates new labeled samples using motifs from a pre-defined vocabulary, enabling connections between tasks through these augmented motifs.

\stitle{Summary.}
Few-shot graph classification presents the significant challenge of \emph{effectively transferring knowledge from base graph classes to novel graph classes to enhance the performance of the latter}. This challenge has inspired a multitude of research, showcasing diverse strategies to tackle this issue, as summarized in Table~\ref{table.few-shot-graph-class}. These studies not only leverage meta-learning strategies, but also employ additional techniques to bolster performance, such as data augmentation, super-class graph construction, \etc\
Nonetheless, certain specific settings within this critical task still demand further exploration. For instance, in a cross-domain scenario, the transfer of knowledge from base to novel classes may pose a greater challenge; on temporal graphs, the complexity of few-shot classification is further amplified by the temporal patterns. Furthermore, few-shot graph-level classification on HINs could also present an intriguing research direction.

\section{\name\ Problems: Structure Imbalance} \label{sec.problem-structure}

Unlike other data types such as images or text, graph data inherently contains topological structures that may exhibit imbalance. The investigation of structure imbalance in graphs has increasingly drawn the attention of researchers due to its prevalence in real-world scenarios. In this section, we extend our discussion for the taxonomy of Problems, examining structure imbalance in graphs from the perspectives of node (Section~\ref{sec.node-level-structure-imba}), edge (Section~\ref{sec.edge-level-structure-imba}), and graph (Section~\ref{sec.graph-level-structure-imba}) levels, as depicted in the right part of Fig.~\ref{fig.taxonomy-problems}.

\subsection{Node-Level Structure Imbalance} \label{sec.node-level-structure-imba}

Node-level structure imbalance arises when the contextual structures surrounding each node exhibit an unequal distribution. A prominent indicator of node-level structure is node degree, representing the count of neighboring nodes and reflecting the vicinity richness of a node. In this section, we explore node-level structure imbalance from two perspectives: imbalanced node degrees and topologies, as summarized in Table~\ref{table.imba-node-degree}. These aspects have significant implications for various real-world applications, such as cold-start recommendations, and knowledge graph enrichment.

\subsubsection{Imbalanced Node Degrees} \label{sec.imba-node-degree}

In graphs, node degrees often follow a long-tail distribution, with \emph{head nodes}---those with high degrees---benefiting from richer structural information and thus achieving superior performance in downstream tasks like node classification \cite{liu2020towards,liu2021tail}. Conversely, \emph{tail nodes} with low degrees have limited topological information, hindering their performance. Research typically targets two main tasks: improving the performance of tail nodes, and addressing the more challenging \emph{cold-start nodes} that are isolated in the graph.
These tasks find applications in various domains, such as predicting new papers in a citation network or new items in an e-commerce network.

\begin{table*}[!t]
\centering
\small
\addtolength{\tabcolsep}{0pt}
 \resizebox{1\linewidth}{!}{
\begin{NiceTabular}{@{}l||c|c|c|l@{}}
\toprule
 \multicolumn{1}{c||}{\textbf{Tasks}}	&    \textbf{Degree-aware modulation} 	& \textbf{Meta-learning} 	&   \textbf{Knowledge distillation}  &  \multicolumn{1}{c}{\textbf{Other techniques}} \\ \midrule\midrule
 \multirow{3}{*}{Tail node embedding}   & \multirow{3}{*}{\cite{wu2019net,tang2020investigating}}  &  \multirow{3}{*}{\cite{liu2020towards,hao2021pre,yang2022few}}  & \multirow{3}{*}{-} & Neighborhood translation \cite{liu2021tail}  \\
  &  &  &  & Hybrid-order proximities \cite{niu2020dual}  \\
 &  &  &  & Reweighting \cite{kojaku2021residual2vec,xia2022cengcn,virinchi2023blade} \\\midrule
 Cold-start node embedding   & -  &  -  &    \cite{zheng2022cold}  & \multicolumn{1}{c}{-}      \\ \midrule\midrule
 \multirow{2}{*}{Node topology imbalance}  & \multirow{2}{*}{-} &  \multirow{2}{*}{-}  & \multirow{2}{*}{-} & Reweighting \cite{chen2021topology,sun2022position}      \\
   &  &    &  & Graph geometric embedding \cite{fu2023hyperbolic}      \\\midrule\midrule
   \multirow{2}{*}{Long-tail entity embedding on KGs}  & \multirow{2}{*}{\cite{zeng2020degree}} &  \multirow{2}{*}{\cite{baek2020learning,MaKEr,chen2022meta,chen2023meta}}  & \multirow{2}{*}{-} & Open knowledge enrichment \cite{cao2020open}      \\
   &  &    &  & Synthetic data generation \cite{shomer2023toward}      \\\bottomrule
\end{NiceTabular}}
\caption{Summary of node-level structure imbalance.} \label{table.imba-node-degree}
\end{table*}

\stitle{Tail node embedding.}
Tail node embedding seeks to boost performance for tail nodes, which often underperform compared to head nodes. 
To tackle this issue, Demo-Net \cite{wu2019net} and SL-DSGCN \cite{tang2020investigating} utilize degree-specific GNNs to capture unique structural patterns across nodes of different degrees.
Contrastingly, meta-tail2vec \cite{liu2020towards} and Tail-GNN \cite{liu2021tail} implement knowledge transfer mechanisms. The former leverages meta-learning to transfer knowledge from head to tail nodes, redefining tail node embedding as a few-shot embedding regression task. Meanwhile, the latter proposes a neighborhood translation technique to bridge the head and tail nodes for knowledge transfer.

Degree imbalance has also been explored in various other tasks.
For instance, Residual2Vec \cite{kojaku2021residual2vec} delves into the task of fair path sampling, where imbalance may stem from the differing degrees of nodes. \del{To achieve fairness, it reweighs the loss function based on both node labels and degrees.}
In contrast, CenGCN \cite{xia2022cengcn} addresses the issue of dominance by hub vertices in scale-free networks, where these vertices can disproportionately propagate influential information due to vertex imbalance. \del{Notably, CenGCN quantifies the similarity between hub vertices and their neighbors, and applies graph transformations through edge weight adjustments and self-connections, effectively mitigating this issue.}
Meanwhile, BLADE \cite{virinchi2023blade} confronts the degree discrepancy problem in link prediction by using biased neighborhood sampling. \del{By creating neighborhoods of varying sizes based on node connectivity, it optimizes performance across different node degrees.}

In \emph{e-commerce networks}, tail node embedding aims to enhance the performance of less-frequent users and shops. 
One typical work DHGAT \cite{niu2020dual} achieves this by leveraging first- and second-order proximities of user histories, utilizing attentive mechanisms to further exploit the neighbors of target entities. Hao \etal\ \cite{hao2021pre}, drawing from the concept of meta-tail2vec, apply regression to tail nodes to facilitate knowledge transfer, enabling cold-start recommendations.

In \emph{dynamic networks}, newly arriving nodes typically confront structural limitations. To address this, MetaDyGNN \cite{yang2022few} targets few-shot link prediction. It harnesses a meta-learning approach to create a hierarchical architecture that applies interval- and node-wise adaptations, thereby facilitating the knowledge transfer from existing to new nodes. 
\del{The approach also includes a dynamic GNN model to efficiently exploit local node structures.}

\stitle{Cold-start node embedding.}
In certain extreme cases, graphs may contain numerous isolated nodes, known as cold-start nodes, which have no neighbors and thus pose a challenge for conventional graph representation learning approaches. To tackle this issue, techniques typically employ the knowledge distillation paradigm \cite{gou2021knowledge}. These approaches utilize a teacher network for knowledge extraction on information-sufficient nodes and a student network for knowledge application on cold-start nodes, ultimately improving the performance of the latter.

Cold Brew \cite{zheng2022cold} is a notable approach designed to address cold-start node representation learning by leveraging the concept of knowledge distillation. Specifically, the method employs a GNN network as the teacher network to extract knowledge from information-sufficient nodes and a multilayer perceptron (MLP) as the student network to distill the knowledge from the teacher. Consequently, the student network can predict suitable representations for cold-start nodes without relying on neighborhood information.

\stitle{Summary.}
The central challenge in managing imbalanced node degrees lies in the \emph{efficient knowledge transfer from head nodes to tail or cold-start nodes}. Various strategies have emerged to address this, as summarized in Table~\ref{table.imba-node-degree}. Notably, degree-aware model modulation and meta-learning are frequently utilized for tail node embedding, while knowledge distillation is favored for cold-start nodes. For further details of these techniques, we direct readers to Section~\ref{sec.knowledge-transfer}. Certain areas still warrant further exploration. For instance, there is a notable dearth of research on the pivotal task of cold-start node embedding. Moreover, while knowledge distillation is typically utilized for cold-start node embedding, its potential for handling other similar tasks, such as tail node embedding, could also be investigated.
Furthermore, we will discuss long-tail entity embedding on KGs, a related task, in Section~\ref{sec.tail-entity-emb}.

\subsubsection{Node Topology Imbalance} \label{sec.node-topo-imba}

In contrast to the frequently studied class imbalance rooted in the quantity-centered disparity in the number of labeled samples across classes (\ie, imbalanced node classification as discussed in Section~\ref{sec.imba-node-class}), \emph{topology imbalance} on graphs shifts the focus to the positional distribution of labeled samples. This positional discrepancy critically influences the label propagation process on the graph. Ideally, in label propagation, the influence boundaries of labeled nodes should coincide with the true class boundaries. However, the specific positions of labeled nodes can trigger a displacement of these influence boundaries away from the true class boundaries, resulting in skewed model decision boundaries. Classes with labeled nodes having influence boundaries more closely aligned with the true class boundaries tend to propagate label information more effectively than those without such alignment, leading to an imbalance, as illustrated in Table~\ref{table.graph-imba-issues}. This uneven distribution could potentially induce biases in learning representations, thereby impacting the performance of nodes across different classes \cite{chen2021topology}.

To address this, ReNode \cite{chen2021topology} reweights labeled nodes based on their proximity to class boundaries, improving performance particularly for boundary-near and remote nodes. 
They also develop a metric to quantify this imbalance using influence conflict detection.
Another work PASTEL \cite{sun2022position} combats topology imbalance by optimizing information propagation paths. It aims to alleviate under-reaching and over-squashing effects by enhancing intra-class connectivity and employing a position encoding mechanism. 
\del{PASTEL also uses a class-wise conflict measure for edge weights to facilitate node class separation.}

In graph data with hierarchical structures, \emph{hierarchy imbalance} represents an unequal distribution of labeled nodes across hierarchical levels. This issue may impede the effective learning and classification of nodes in underrepresented hierarchy levels, also affecting the decision boundary of the classifier \cite{fu2023hyperbolic}.
To tackle this challenge, HyperIMBA \cite{fu2023hyperbolic} utilizes hyperbolic geometric embedding to gauge the hierarchy of labeled nodes. Subsequently, it adjusts label information propagation and alters the objective margin based on the hierarchy of the node, thereby addressing the issues caused by hierarchy imbalance.

\stitle{Summary.}
In addition to the extensively studied imbalance issues (\eg, class or degree imbalance) that affect the performance of node classification, the topology of nodes, which depicts the interconnectedness of nodes from the label information, remains underexplored. Specifically, given the importance of label propagation on graphs, the connecting topologies could play a pivotal role in the node classification task, indicating a need for further investigation in this area.

\subsubsection{Long-Tail Entity Embedding on KGs} \label{sec.tail-entity-emb}

Knowledge graphs often present a long-tail distribution of triplets across entities, potentially causing performance bias \cite{farid2016lonlies} in applications such as KG enrichment. To address this issue, a series of approaches have been proposed.

GEN \cite{baek2020learning} is a transductive meta-learning framework predicting links for new entities. It uses aggregation on neighboring entities and relations, and employs MAML \cite{finn2017model} for knowledge transfer from seen (head) to unseen (tail) entities.
The works of MaKEr \cite{MaKEr}, MTKGE \cite{chen2023meta}, and MorsE \cite{chen2022meta} follow a similar paradigm, employing both meta-learning and GNN encoding to tackle this issue. Specifically, MaKEr operates within a federated setting, basing its process on representations calculated on their constructed relation position graph. MTKGE, on the other hand, concentrates on the relative position and temporal sequence patterns between relations to extrapolate missing facts in emerging temporal knowledge graphs. MorsE, meanwhile, focuses on inductive knowledge graph embedding, by incorporating two key modules---an entity initializer and a GNN modulator---for enhanced entity embedding.
In addition, KG-Mixup \cite{shomer2023toward} generates synthetic triples with Mixup for data augmentation, aiming to enhance KG completion performance, particularly on tail entities.

Some approaches also leverage additional knowledge to address this issue. 
For instance, Zeng \etal\ \cite{zeng2020degree} propose a degree-aware co-attention network to guide information fusion from different sources for tail entity alignment, forming an iterative paradigm to improve KG completion.
Another work OKELE \cite{cao2020open} aims to increase tail entity triplets by extracting open knowledge from the Web. \del{It constructs an entity-property graph on the given KG for attentive GNN-based representation learning.}

\stitle{Summary.}
Tail entities on KGs share similar characteristics with tail nodes on graphs, as both have a scarcity of linked neighbors. Notably, some similar techniques, such as meta-learning, are employed to address tail entity embedding. Moreover, supplementary knowledge, such as auxiliary KGs, serves as an additional source of information to aid in resolving this issue. Consequently, the solutions proposed for tail entity or node embedding can draw inspiration from each other to enhance their respective performance.

\subsection{Edge-Level Structure Imbalance} \label{sec.edge-level-structure-imba}

Edge-level structure imbalance in complex graphs, like KGs, manifests as varying edge (relation) frequencies, often following a long-tail distribution \cite{xiong2018one,gao2019hybrid}. This imbalance can bias the model, marginalizing long-tail relations and leading to suboptimal performance---a challenge that is garnered significant research attention due to its implications in real-world scenarios. We delve into edge-level topology imbalance through three lenses: few-shot relation classification, zero-shot relation classification, and reasoning with few-shot relations, and summarize them in Table~\ref{table.edge-structure-imba}. Note that the frequency of edges reflects their structural distribution in KGs, hence we classify this aspect as structure imbalance. Alternatively, it can also be viewed as class imbalance, considering edges as labeled instances of the edge class.

\begin{table*}[!t] 
    \centering
    \small
    \addtolength{\tabcolsep}{-1.5pt}
   \resizebox{1\linewidth}{!}{%
    \begin{NiceTabular}{@{}l||c|c|p{7cm}@{}}
    \toprule
  \Block[c]{2-1}{\textbf{Tasks}} & \multicolumn{2}{c|}{\textbf{Meta-learning techniques}} & \Block[c]{2-1}{\textbf{Other techniques}}\\ \cmidrule{2-3}
  & Optimization-based & Metric-based &  \\\midrule\midrule
    \makecell[l]{Few-shot relation classification} & \makecell[c]{\cite{chen2019meta,niu2021relational,zhenzhen2022improving,wu2023hierarchical,zheng2022subgraph}} & \makecell[c]{\cite{xiong2018one,gao2019hybrid,zhang2020few,han2021multi,zhang2021knowledge,xiao2021hmnet,sheng2020adaptive,dou2022function,jiang2021metap,zhang2022multiform,ren2022granularity,xiao2021adaptive,li2022learning}} & \Block[l]{1-1}{Neural snowball \cite{gao2020neural}, error mitigation \cite{wang2021reform}, data augmentation \cite{wang2019tackling}, gaussian metric learning \cite{zhang2021gaussian}, path-based interactions \cite{xu2021pint}, entity interactions \cite{li2022learning}, neural process \cite{luo2023normalizing}}  \\ \midrule
    Zero-shot relation classification & - & \cite{gong2021zero} & \multicolumn{1}{c}{GAN \cite{qin2020generative}}    \\ \midrule
    Few-shot multi-hop reasoning on KGs & \cite{lv2019adapting,zhang2020few,zheng2021hardness,zhang2022adapting}  & - & \multicolumn{1}{c}{-} \\\bottomrule
     \end{NiceTabular}}
     \caption{Summary of edge-level structure imbalance.}
      \label{table.edge-structure-imba}%
\end{table*}

\subsubsection{Few-Shot Relation Classification}
In scenarios such as KG enrichment, there are often \emph{novel relations} that have only a few available triplets. In contrast, numerous \emph{base relations} are provided to support classification tasks involving these novel relations. In this context, the objective of Few-Shot Relation Classification (FSRC) is to classify triplets into novel relations, with only a limited number of triplets serving as supervision.

\stitle{Generic FSRC.}
Within the realm of FSRC on KGs using \emph{metric-based meta-learning} methods \cite{snell2017prototypical,vinyals2016matching}, various studies exhibit similarities in approach, albeit with distinct contributions. Certain investigations implement unique techniques for metric computation and encoding. For instance, Gmatching \cite{xiong2018one} and FSRL \cite{zhang2020few} employ matching networks, differing in their methods of entity representation. While Gmatching utilizes a GNN encoder, FSRL infers entity embeddings by encoding heterogeneous neighbors. Furthermore, MetaP \cite{jiang2021metap} exploits co-occurrence relation patterns in meta-tasks to enhance the effectiveness of proximity calculations within matching networks.

A common approach revolves around the usage of attention mechanisms and prototypical networks. For instance, Gao \etal\ \cite{gao2019hybrid} and KEFDA \cite{zhang2021knowledge} utilize attention-based prototypical networks. The former \cite{gao2019hybrid} is distinguished by its instance- and feature-level hybrid attention mechanism, enhancing prototype calculation accuracy. KEFDA \cite{zhang2021knowledge}, on the other hand, stands out by incorporating general and domain knowledge into the model.
The theme of attention mechanisms continues in IAN \cite{han2021multi} and HMNet \cite{xiao2021hmnet}, with each focusing on different aspects. IAN \cite{han2021multi} computes inter- and intra-instance correlations to guide fusion operations, while HMNet \cite{xiao2021hmnet} conducts dual scoring from both entity and relation perspectives for effective link prediction.

A number of studies present other contributions based on prototypical networks.
Both FAAN \cite{sheng2020adaptive} and work \cite{xiao2021adaptive} utilize adaptive learning models to tackle FSRC. Specifically, FAAN introduces an adaptive attentional network to capture dynamic properties, while the latter proposes an adaptive mixture mechanism that incorporates label words into class prototypes. On the other hand, MULTIFORM \cite{zhang2022multiform} enriches entity representations by incorporating multi-modal contexts. And another research \cite{ren2022granularity} adopts a granularity-aware method, depicting each relation as an area to account for variations in granularities.

Among studies employing \emph{optimization-based meta-learning} methods for FSRC on KGs, MAML \cite{finn2017model} serves as a key foundation. Both MetaR \cite{chen2019meta} and MTransH \cite{niu2021relational} employ MAML for link prediction and few-shot training respectively, with MetaR using an MLP for relation representation and MTransH integrating a gated and attentive neighbor aggregator.
Li \etal\ \cite{zhenzhen2022improving} deviate by jointly training a prototype and instance encoder, enhancing the relation classification accuracy of the model. In addition, Mick \cite{geng2020mick}, HiRe \cite{wu2023hierarchical} and Meta-iKG \cite{zheng2022subgraph} explore different facets: Mick explores FSRC from the perspective of limiting available data during training, HiRe focuses on hierarchical relational learning across different levels, while Meta-iKG utilizes local subgraphs for efficient pattern learning and generalization across few-shot and large-shot relations.

Some studies highlight \emph{other techniques} or \emph{aspects of KGs} for FSRC.
For instance, Gao \etal\ \cite{gao2020neural} deploy an iterative neural snowball process with fake labels for novel relations, relying on a relational Siamese network \cite{koch2015siamese} and a relation classifier.
REFORM \cite{wang2021reform} uses an error mitigation mechanism alongside attentive neighborhood encoding and cross-relation aggregation for error-aware completion. 
In contrast, P-INT \cite{xu2021pint} encodes relations via expressive paths for accurate matching. Similar to IAN \cite{han2021multi}, CIAN \cite{li2022learning} leverages attention mechanisms to capture both intra- and inter-entity interactions for better entity pair representations.
Meanwhile, NP-FKGC \cite{luo2023normalizing} merges normalizing flows and neural processes to manage complex relations and estimate uncertainties, aided by an attentive relation path-based GNN to better capture KG path information.

\stitle{FSRC with uncommon entities.}
Wang \etal \cite{wang2019tackling} consider an extreme setting with uncommon entities and long-tailed relations. They use textual descriptions (\ie, description encoder) to extract crucial information for entity and relation embeddings. To address data limitations, they propose a triplet generator to synthesize fake triplets for augmentation, augmenting both limited entities and relations. 
Reptile \cite{nichol2018first} is employed for model optimization.

\stitle{FSRC on uncertain KGs.}
In uncertain KGs, where each triplet has a confidence score indicating its certainty, few-shot relation classification remains a challenge. Zhang \etal\ address this issue with Gaussian Metric Learning \cite{zhang2021gaussian}, which completes missing facts and confidence scores with few available examples. They propose a Gaussian neighbor encoder to represent relation facts as multi-dimensional Gaussian distributions, simultaneously learning semantic features and internal uncertainty. Additionally, they propose a Gaussian matching function considering fact qualities to discover new facts and predict confidence scores.

\stitle{Few-shot inverse relation classification.}
For few-shot inverse relation classification, FAEA \cite{dou2022function} employs a hybrid attention model to attend class-related words based on meta-learning. It leverages function-words enhanced attention to effectively compute representations for both support and query instances, facilitating message passing and similarity calculation between queries and prototypes.

\stitle{Summary.}
The principal challenge of few-shot relation classification is \emph{the effective transfer of knowledge from data-rich relations to long-tail relations}, aiming to bolster the performance of the latter. As depicted in Table~\ref{table.edge-structure-imba}, most existing approaches rely heavily on meta-learning techniques, such as prototypical networks and MAML, to counter the imbalance issue. Some methods also employ different techniques like the neural snowball \cite{gao2020neural}, error mitigation \cite{wang2021reform}, among others. Moreover, several extreme scenarios have also been explored, including few-shot relation classification with uncommon entities, within uncertain knowledge graphs, and few-shot inverse relation classification. 
Delving deeper into these extreme scenarios could further advance progress in FSRC.
Additionally, benchmarks and leaderboards, such as FewRel\footnote{\url{https://github.com/thunlp/FewRel}.} \cite{han2018fewrel,gao2019fewrel}, have been introduced for comparison.

\subsubsection{Zero-Shot Relation Classification}

Zero-shot relation classification on KGs, classifying triplets without given supervision, typically leverages auxiliary text descriptions for zero-shot relation representation.
For instance, ZSGAN \cite{qin2020generative} employs GAN \cite{goodfellow2014generative} to transfer knowledge from labeled to zero-shot relations. The generator aims to produce zero-shot relation embeddings using text descriptions, while the discriminator classifies the generated embeddings into respective classes.
Putting it differently, ZSLRC \cite{gong2021zero} utilizes side information through hypernym and keyword extraction for new relation type detection. It then implements a side-information-enhanced prototypical network for few-shot relation classification by computing a weighted side information embedding for each relation.

\subsubsection{Few-Shot Multi-Hop Reasoning on KGs}

Multi-hop reasoning on KGs offers an effective method for inferencing target entities given query entities and relations, a technique crucial for tasks like query answering. However, it is often challenged by a limited number of labeled triplets, leading to a few-shot setting.

Efforts to tackle this issue, such as Meta-KGR \cite{lv2019adapting} and FIRE \cite{zhang2020fewmulti}, usually utilize reinforcement learning \cite{williams1992simple} for multi-hop reasoning and MAML to extract meta-knowledge to address the few-shot problem.
To improve the generalization abilities, THML \cite{zheng2021hardness} further introduces a hardness-aware meta-reinforcement learning method. This method trains hardness-aware batches using a two-level hardness-aware sampling approach, addressing low reasoning accuracies over challenging relations.
In a similar vein, ADK-KG \cite{zhang2022adapting} enhances the standard reinforcement learning and MAML modules with text-enhanced heterogeneous GNN for better node embeddings. It also incorporates a knowledge distillation module using unlabeled data to generate synthetic labels, further improving model training.

\subsection{Graph-Level Structure Imbalance} \label{sec.graph-level-structure-imba}

The intricate interconnections within graphs can lead to structural imbalance across graphs. This type of imbalance often manifests as differences in graph sizes \cite{liu2022size}, or topology groups \cite{zhao2022topoimb}, as illustrated in Table~\ref{table.graph-imba-issues}. Typically, graphs with advantageous structures, such as larger sizes, are more expressive and subsequently yield superior performance compared to their counterparts, which may introduce bias in applications such as molecular or protein prediction.

\subsubsection{Imbalanced Graph Sizes}

Graphs with larger sizes (\eg, number of nodes), which offer more intricate structures, are generally more expressive than smaller ones. This disparity in size often results in performance biases in graph-level tasks such as graph classification \cite{liu2022size}.
To address this challenge, SOLT-GNN \cite{liu2022size} is designed to enhance the performance of smaller, or ``tail'', graphs. SOLT-GNN first identifies co-occurrence patterns in the structures of larger, or ``head'', graphs, to generate transferable knowledge. This information is then utilized to augment tail graphs by predicting co-occurrence patterns that supplement their existing structures. 
Thus, the performance of tail graphs can be improved accordingly.

\subsubsection{Imbalanced Topology Groups}
Contrasting with the node topology imbalance discussed in Section~\ref{sec.node-topo-imba}, \emph{imbalanced topology groups} \cite{zhao2022topoimb} refer to the unequal distribution of topology groups, such as topological motifs, within individual classes in graph data. Take, for example, the Mutag dataset \cite{zhao2022topoimb}, where molecular graphs of the Mutagenic class contain two distinct topology groups: one associated with the motif $NO_2$ and the other with $NH_2$. In certain scenarios, the $NO_2$ group may substantially outnumber the $NH_2$ group, which results in the $NO_2$ motif having stronger associations with the class than the less frequent $NH_2$ motif. This imbalance, anchored in motif distribution, can lead to a paucity of training instances for minority groups. Consequently, classifiers may overfit to the majority topology (motif) groups, hampering effective learning and classification of instances in the underrepresented minority topology groups.
To address this issue, TopoImb \cite{zhao2022topoimb} dynamically updates the discovery of topology groups and assigns importance weights to under-represented instances during training, by incorporating a topology extractor and a training modulator. This approach improves the effectiveness of learning on minority topology groups and addresses the issue of over-fitting to majority groups.

\stitle{Summary}
In conclusion, the problem of graph-level structure imbalance is still a largely underexplored domain that calls for more in-depth study. The core challenge in this area is \emph{the enhancement of the expressiveness of tail (sub)graphs or groups}. Predominantly, current methods tend to utilize knowledge transfer or reweighting techniques to address this issue. By drawing inspiration from the tasks previously discussed, a variety of strategies could potentially be adapted and applied to grapple with this intricate yet under-studied task. Additionally, other types of graph-level imbalance issues that could potentially impact the performance of graph learning models may also require investigation, further emphasizing the need for continued research.

\section{Techniques of \name} \label{sec.techniques}

As detailed in Section \ref{sec.introduction}, graph imbalance issues typically involve two primary components: the high- and low-resource parts. This leads to two main task categories: \emph{improving the low-resource part} and \emph{balancing both high and low-resource parts}. In this case, Section~\ref{sec.impr-low-part} will elaborate on tasks aimed at improving the performance of the low-resource part, such as few-shot node classification \cite{zhou2019meta,ding2020graph,liu2021relative} and tail/cold-start node representation learning \cite{liu2021tail,zheng2022cold}. In addition, Section~\ref{sec.balan-both-parts} will explore tasks aiming to equalize performance across both parts, essential in scenarios like imbalanced classification tasks \cite{shi2020multi,zhao2021graphsmote,qu2021imgagn}.

\begin{figure}[t]
\centering
\includegraphics[width=0.99\linewidth]{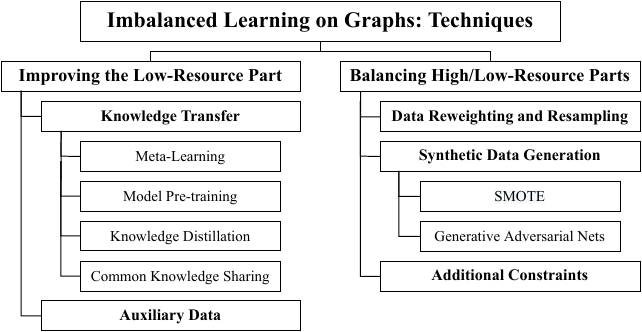}
\caption{Taxonomy of Techniques.}
\label{fig.taxonomy-techniques}
\end{figure}

Our taxonomy, as summarized in Fig.~\ref{fig.taxonomy-techniques}, offers a well-organized and extensive overview of the existing literature across multiple tasks. 
This classification approach not only effectively differentiates various scenarios, but it also aids in identifying strategies bespoke to these specific settings, thereby facilitating effective management of graph imbalance issues, which we will discuss further in Section~\ref{sec.tech-choose}.

We also summarize the literature \wrt this taxonomy in Table~\ref{table.techniques-taxonomy}. 
It is important to note that this categorization is not inflexible: some studies may incorporate techniques from another branch to mitigate imbalance issues in their own domains. For example, some studies \cite{shomer2023toward} apply Mixup for synthetic data generation, enhancing the performance of the low-resource part. Others, such as \cite{wang2022imbalanced,yun2022lte4g}, use techniques like common knowledge sharing or knowledge distillation for knowledge transfer, seeking balanced performance across both parts. Note that, despite these cross-branch applications, our discussion primarily centers on major strategies for tackling imbalance.

\begin{table*}[!t]
    \centering
   \small
    \addtolength{\tabcolsep}{-1mm}
    \resizebox{1.0\linewidth}{!}{
    \begin{NiceTabular}{@{}c||c||c|c|p{9cm}@{}} 
    \toprule
     & \multicolumn{1}{c||}{\textbf{Techniques}} & \multicolumn{3}{c}{\textbf{Literature}}  \\ \midrule\midrule
     \Block[c]{7-1}{\makecell*[c]{Improving \\ the \\ low- \\ resource \\ part}} & \Block[c]{6-1}{Knowledge transfer} & \Block[c]{2-1}{Meta-learning} & \Block[c]{1-1}{Optimization-based} & \Block[c]{1-1}{\cite{zhou2019meta,lv2019adapting,chen2019meta,ma2020adaptive,wang2020graph,geng2020mick,liu2020towards,zhang2020fewmulti,baek2020learning,liu2021relative,zheng2021hardness,wang2021property,niu2021relational,guo2021few,zhang2022fewheterogeneous,fu2022meta,zhang2022adapting,zhang2022hg,liu2022few,yang2022few,wang2019tackling,ding2021few,qian2021distilling,zhou2022task,lu2022spatio,borde2022graph,zhu2023few,wu2023hierarchical,meng2023meta,zheng2022subgraph,wangfewshotnode2023}}  \\ 
     \cmidrule{4-5} 
     & & & \Block[c]{1-1}{Metric-based} & \Block[c]{1-1}{\cite{gao2019hybrid,yao2020graph,ding2020graph,huang2020graph,lan2020node,gong2021zero,zhang2021knowledge,hassani2022cross,wang2022faith,dou2022function,lu2022geometer,wang2022task,tan2022graph,xiong2018one,zhang2020few,sheng2020adaptive,han2021multi,zhang2021gaussian,xiao2021hmnet,jiang2021metap,crisostomi2022metric,ren2022granularity,zhang2022mul,xiao2021adaptive,zhang2022multiform,zhenzhen2022improving,li2022learning}}  \\ 
     \cmidrule{3-5}
     &  & Model pre-training & \multicolumn{2}{c}{GNN parameters transfer \cite{ijcai2022p500}, contrastive learning \cite{tan2022supervised,tan2022transductive}, prompting \cite{liu2023graphprompt}}  \\ \cmidrule{3-5}
     &  & Knowledge distillation & \multicolumn{2}{c}{GNNs to MLPs \cite{zheng2022cold}, KG models to MLPs \cite{qin2020generative}, Random knowledge distillation \cite{ma2022deep}} \\ \cmidrule{3-5}
     &  & \Block[c]{2-1}{\makecell*[c]{Common \\ knowledge sharing}} & Data sharing & \Block[c]{1-1}{Super-classes \cite{Chauhan2020FEWSHOT}} \\ \cmidrule{4-5}
     & & & Model sharing & \Block[c]{1-1}{\cite{wu2019net,tang2020investigating,liu2021tail,liu2022size}} \\ \cmidrule{2-5}
     & \multirow{1}{*}{Auxiliary data} &  \multicolumn{3}{c}{Alignment data \cite{zeng2020degree}, auxiliary descriptions \cite{gong2021zero,wang2021zero,yue2022dual}} \\ \midrule\midrule
    \Block[c]{5-1}{\makecell*[c]{Balancing \\ high/low- \\ resource \\ parts}} & \multirow{2}{*}{Reweighting and resampling} & \multicolumn{3}{p{15.5cm}}{Reweighting \cite{chen2021topology,tang2022rethinking,chen2022antibenford,ying2016pfraudetector,zhao2022topoimb,xia2022cengcn,fu2023hyperbolic,liu2022dagad,zhang2021fraudre,wang2021distance}, resampling \cite{cui2022allie,dong2022bi,huang2022auc,shi2022h2,wang2017gang,li2022devil,lu2022bright,dou2020enhancing,li2022devil,liu2020alleviating,liu2021pick,zheng2019addgraph,ren2021ensemfdet,sun2022position,virinchi2023blade} 
    } \\\cmidrule{2-5}
    & \Block[c]{3-1}{\makecell*[c]{Synthetic data \\ generation}} &  
     SMOTE & \multicolumn{2}{c}{SMOTE \cite{zhao2021graphsmote}, Mixup \cite{park2022graphens,wu2022graphmixup,liu2023imbalanced}} \\ \cmidrule{3-5}
     &  & 
     GAN & \multicolumn{2}{c}{\cite{qu2021imgagn,chen2020generative,ding2021inductive,deng2022graph,bhatia2021graphanogan}} \\ \cmidrule{3-5}
     &  & 
     Other methods & \multicolumn{2}{c}{Predictive data generation \cite{zhu2020heterogeneous,zhao2021action}, label generation \cite{mathew2021defraudnet,tian2023sad,li2021live}}
     \\ \cmidrule{2-5}
     & \multirow{3}{*}{Additional constraints} &  \multicolumn{3}{p{15.5cm}}{Condition relax constraints \cite{wang2018rsdne,wang2020network,wang2021expanding,xu2021towards}, imbalance constraints \cite{song2022tam,zhao2021synergistic}, class separation constraints \cite{shi2020multi,deng2021graph,desai2021graph,wang2022calibrated,qi2022mad,chen2022deep,zhang2022reconstruction,guo2022subset,huang2022hop,zhou2022unseen,chen2022graphad,bhatia2020midas,ding2021cross,liu2022deep,luo2022comga,rashidi2016node,teng2017anomaly,liu2017accelerated,zhang2017hidden,hooi2017graph,bojchevski2018bayesian,liang2018semi,bandyopadhyay2019outlier,wu2019uncovering,qin2022explainable,cui2022meta,zhang2022unsupervised,zhou2018sparc,bandyopadhyay2020integrating,ahmed2021graph,yuan2017spectrum,zhuang2023subgraph,gao2023addressing,huang2023unsupervised,zhao2020error,peng2020deep,jin2021anemone,goodge2022lunar,zhou2021subtractive,liu2021anomaly,cheng2020graph,li2019specae,ding2019deep,li2017radar,yang2019mining,aggarwal2021signature,nguyen2023detecting,wang2017gang,zhu2020mixedad,ding2019interactive,gutierrez2020multi,peng2018anomalous,hu2016embedding,dong2022bi,huang2022auc,gao2023alleviating,hu2021turbo,tang2022rethinking,shi2022h2,wang2021modeling,zhang2020gcn,wang2021decoupling,fan2018gotcha,yoon2021graph,ranjbar2018qanet,teng2018deep,wang2021bipartite,chen2020anomaly,ofori2021topological,yu2018netwalk,lin2019fraud,fond2018designing,yoon2019fast,liu2017holoscope,liu2018contrast,sharma2018nhad,qiu2022raising,perozzi2018discovering,wu2017query,sun2020anomaly,chen2022antibenford,zhu2022binarizedattack,hou2023efficient,zambon2018concept,zhao2020multivariate,manzoor2016fast,duan2020aane,cao2017hitfraud,miz2019anomaly,cai2021structural,chang2021f,wang2019nodes,liu2021intention}} \\
    \bottomrule
     \end{NiceTabular}}
     \caption{Literature categorization of imbalanced learning on graphs \wrt the taxonomy of techniques.} \label{table.techniques-taxonomy}
\end{table*}

\subsection{Improving the Low-Resource Part} \label{sec.impr-low-part}

The primary goal of tasks designed to improve the low-resource part is to boost their performance, despite having less data compared to the high-resource part. They commonly exploit the high-resource part, which often holds a wealth of knowledge, as a resource to enrich the low-resource part. This category includes various graph tasks such as few-/zero-shot node/edge/graph classification, few-shot link prediction, or few-shot reasoning on KGs, and cold-start/tail node/entity/graph embedding, among others. To handle the imbalance issues, two main techniques are typically used: \emph{knowledge transfer} and the incorporation of \emph{auxiliary data}, in order to leverage existing resources to supplement low-resource parts.

\subsubsection{Knowledge Transfer} \label{sec.knowledge-transfer}

In typical settings, there often exists a data part that is characterized by a wealth of source knowledge, such as the base classes within the framework of few-shot node classification, or high-degree nodes in tail node embedding situations. Knowledge transfer intends to harness this reservoir of knowledge residing within the data-rich source, and transfer it to the knowledge-deficient segment, thus bolstering its performance. To enable this, a suite of techniques has been engineered to facilitate efficient knowledge transfer between the source and target data, thereby enhancing the efficacy of imbalanced learning on graphs.

\stitle{Meta-learning.}
Meta-learning, distinct from traditional machine learning methods, extracts versatile meta-knowledge from a series of base meta-tasks with a shared distribution. This knowledge is then applied to novel meta-tasks to enhance predictive performance. Many studies on \name\ have utilized this approach to extract transferable meta-knowledge from high-resource parts (base meta-tasks) and apply it to low-resource parts (novel meta-tasks). 
Techniques employed often include optimization-based approaches (\eg, MAML \cite{finn2017model}) and metric-based approaches (\eg, prototypical networks \cite{snell2017prototypical}), among others.

\del{MAML, compatible with any optimization-based models, learns a base task prior that allows quick adaptation to the support set of a meta-task, facilitating rapid predictions on the query set. Its effectiveness supports various approaches addressing imbalanced learning on graphs \cite{zhou2019meta,wang2020graph,liu2021relative,zhang2022few,zhang2022hg,ma2020adaptive}.}

\del{Conversely, within a meta-task, prototypical networks aim to identify the center of each class based on the support set, using distance-based metrics to classify novel instances in the query set. This simple but effective method is widely used in imbalanced learning on graphs to leverage transferable knowledge \cite{yao2020graph,ding2020graph,huang2020graph,wang2022task,liu2022few,lan2020node}. Additionally, techniques like Matching Networks \cite{vinyals2016matching} and Siamese Networks \cite{koch2015siamese} are also commonly used in this context.}

\stitle{Model pre-training.} 
The  paradigm of ``pre-training, fine-tuning'' is a widely accepted two-step model training strategy in machine learning. Initially, models are pre-trained on a large dataset using self-supervised learning, which uncovers the underlying knowledge embedded within the data. Following this, the pre-trained models are fine-tuned on a smaller, task-specific supervised dataset. This strategy benefits the fine-tuning stage by providing useful initializations for the models, particularly useful when dealing with limited labeled data. This strategy has demonstrated success across various tasks, setting benchmark results and establishing itself as a robust paradigm \cite{liu2023pre,li2020oscar,Hu2020Strategies}.

Within the sphere of \name, a typical strategy involves pre-training models on the high-resource parts to distill the embedded knowledge, such as pre-training GNN models in alignment with the label information of the high-resource parts \cite{ijcai2022p500,tan2022supervised}. Subsequently, these pre-trained models undergo fine-tuning with limited supervision from the low-resource parts, thereby facilitating the transfer of knowledge to enhance the performance of these low-resource parts.

\stitle{Knowledge distillation.}
Knowledge distillation \cite{gou2021knowledge} is a technique wherein a smaller ``student'' model learns from a larger ``teacher'' model by mimicking its behavior. The teacher model offers soft targets to guide the student model, thereby transferring its knowledge to the student, aiming to boost efficiency while preserving performance \cite{kim2016sequence,park2019relational}.

Regarding \name, one approach involves constructing a comprehensive teacher model for the high-resource part to extract knowledge. To tackle data scarcity in the low-resource part, a student model is developed, guided by the teacher, facilitating training on limited data. This technique has shown noteworthy usage in situations such as cold-start node embedding \cite{zheng2022cold}, as detailed in Section~\ref{sec.imba-node-degree}.
In summary, knowledge distillation is particularly beneficial in scenarios where distinct models are needed to decode patterns within high- and low-resource parts of the graph.

\stitle{Common knowledge sharing.}
Common knowledge sharing forms the crux of knowledge transfer, establishing a conduit between high- and low-resource parts through shared elements. Some studies propose shared data \cite{Chauhan2020FEWSHOT} or shared sub-models \cite{wu2019net,tang2020investigating,liu2021tail,liu2022size} as bridges for knowledge transfer. These shared components serve as critical resources for models associated with both high- and low-resource parts, facilitating further model development while maintaining a conversation via this shared bridge.

These methods are particularly beneficial in \name. Strategies employing shared data construct overarching relationships for both high- and low-resource parts, like super-classes based on original classes \cite{Chauhan2020FEWSHOT}. Establishing this hierarchical structure enables knowledge transfer from high-resource (\eg, majority classes) to low-resource parts (\eg, minority classes) via shared relationships.
In contrast, studies addressing structure imbalance commonly design a shared model for high- and low-resource parts, further incorporating model modulation to adapt the globally shared model to the unique characteristics of each part \cite{wu2019net,tang2020investigating,liu2021tail,liu2022size}. The shared component, or globally shared model, retains common knowledge for both parts and serves as a bridge for knowledge transfer from high-resource (\eg, high-degree nodes \cite{wu2019net,tang2020investigating,liu2021tail} or large-size graphs \cite{liu2022size}) to low-resource parts (\eg, low-degree nodes or small-size graphs).
This approach is generally effective in scenarios where certain common attributes are shared by both parts, such as high-level features from the data-sharing perspective (\eg, super-classes) or common knowledge which can be preserved by a shared model.

\subsubsection{Auxiliary Data}

Auxiliary data in machine learning refers to supplemental information that enhances the primary dataset, consequently improving model performance. For \name, auxiliary data often underpins the learning process of low-resource parts by contributing additional knowledge \cite{hoffman2016learning,vashishth2018reside}.

Auxiliary data in the realm of \name\ takes various forms, enabling it to support multiple tasks. For zero-shot learning, text descriptions can offer class information for each class, facilitating classification in the absence of label support \cite{gong2021zero,wang2021zero,yue2022dual}. Some research centered on knowledge graph completion exploits auxiliary KGs to align with the target KG, providing extra information for KG completion concerning long-tailed entities \cite{zeng2020degree} or zero-shot relations \cite{qin2020generative,gong2021zero}.
As such, the use of auxiliary data can be contemplated in most scenarios of \name\ where it is available.

\subsection{Balancing the High- and Low-Resource Parts} \label{sec.balan-both-parts}

In contrast to the objective of solely improving the low-resource part, balancing the high- and low-resource parts aims to enhance the performance of the low-resource part while minimizing potential performance degradation in the high-resource part. Typical tasks often involve imbalanced node/graph classification \cite{zhao2021graphsmote,wang2022imbalanced}, anomaly detection \cite{liu2021pick,tang2022rethinking}, \etc\ 
To accomplish this, three primary techniques have been proposed: \emph{data reweighting and resampling}, \emph{synthetic data generation}, and the incorporation of \emph{additional constraints}.

\subsubsection{Data Reweighting and Resampling}

Data resampling and reweighting techniques are commonly employed in imbalanced learning to address class imbalance issues within training data. Class imbalance occurs when there is an unequal distribution of classes, leading to a bias towards the majority class and reduced performance on minority classes. Both data resampling and reweighting aim to rebalance the class distribution, allowing the model to learn from minority class examples more effectively. 
Within the scope of \name, data reweighting and resampling methods are prevalently employed to address class imbalance challenges \cite{chen2021topology,cui2022allie}. These techniques are applicable in numerous scenarios plagued by imbalance issues.

\subsubsection{Synthetic Data Generation}

Synthetic data generation techniques are widely employed in \name\ to address the challenge of class imbalance by generating synthetic data for data rebalance. These techniques offer valuable solutions to achieve balanced performance in scenarios where there is a significant disparity between the high- and low-resource parts of the data. 
Specifically, two commonly used techniques for synthetic data generation are \emph{SMOTE} (Synthetic Minority Over-sampling Technique) \cite{chawla2002smote} and \emph{GAN} (Generative Adversarial Networks) \cite{goodfellow2014generative}.

\stitle{SMOTE.}
SMOTE \cite{chawla2002smote} is a well-known approach utilized in imbalanced learning to combat the class imbalance problem within training data. 
It involves creating synthetic samples by interpolating between existing minority class instances. For each minority class instance, SMOTE selects one or more k-nearest neighbors from the same class and generates synthetic examples by linearly interpolating between the selected instance and its neighbors. This process effectively expands the minority class and introduces additional data points to improve its representation during training. 

Furthermore, Mixup \cite{zhang2018mixup}, a data augmentation technique initially proposed for training deep neural networks, can also be leveraged in imbalanced learning scenarios. Mixup extends the idea of interpolation by generating virtual training examples through linear combinations of instance pairs. As a special case of SMOTE, Mixup can also promot model generalization and reducing overfitting \cite{zhang2018mixup}.

In the context of \name, researchers have explored the adaptation of SMOTE and Mixup to address the class imbalance issue \cite{zhao2021graphsmote,park2022graphens,wu2022graphmixup}. 
These methods are particularly beneficial when there is a need to bolster the low-resource segment with additional instances for data rebalance, for tasks like imbalanced node/graph classification.

\stitle{GAN.}
Generative Adversarial Networks (GANs) \cite{goodfellow2014generative} are a class of machine learning techniques composed of two modules, a generator and a discriminator, which are trained together in an adversarial manner. GANs are known for generating realistic and high-quality data, such as images, that closely resemble the original data distribution.

In the context of \name, some research (\eg, IMGAGN \cite{qu2021imgagn}) has explored the use of GANs to generate synthetic samples for minority classes, functioning similarly to SMOTE. 
These GAN-based techniques can help balance the data distribution and improve the model performance on minority classes in imbalanced graph learning scenarios.

\subsubsection{Additional Constraints}

Additional constraints can be effectively incorporated in \name\ to bridge the gap between high- and low-resource parts, to allow knowledge transfer between them while adhering to predefined constraints, thereby enhancing performance. 
For instance, some studies \cite{wang2018rsdne,wang2020network,wang2021expanding} have attempted to relax the intra-class similarity requirement by allowing nodes with the same labels to reside on the same manifold in the embedding space. 
In particular, the selection and design of these constraints depend on the specific characteristics of the dataset and the learning tasks at hand.

\begin{figure}[t]
\centering
\includegraphics[width=0.99\linewidth]{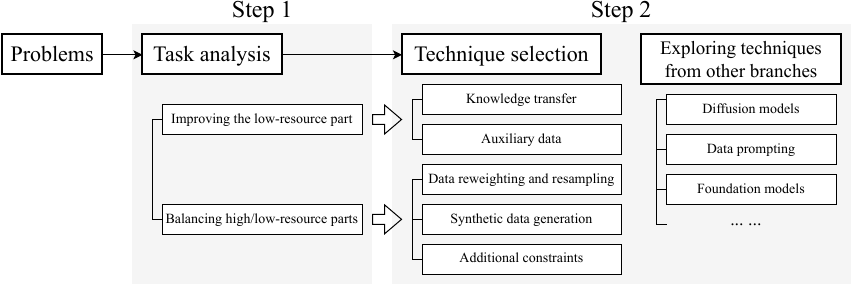}
\caption{Procedure of techniques selection.}
\label{fig.techniques-selection}
\end{figure}

\subsection{Appropriate Techniques Selection} \label{sec.tech-choose}

Selecting the proper technique for \name\ is pivotal, following a specific paradigm, as shown in Fig.~\ref{fig.techniques-selection}. The process comprises task analysis, technique selection, and further exploration of methods from related machine learning fields.

Firstly, \emph{task analysis} involves understanding the data imbalance and its challenges. This step requires determining the data split into high- and low-resource parts and identifying the primary goal: improving the performance of the low-resource part, or balancing both parts. This analysis is the foundation for subsequent technique selection.

Secondly, \emph{technique selection} involves considering various methods to address the identified imbalance issue. Each technique offers a unique perspective and capability to address the problem, making it a potential selection candidate.

For tasks focused on enhancing low-resource part performance, knowledge transfer and the use of auxiliary data are key considerations. Knowledge transfer is particularly beneficial when ample base data is available. Meta-learning is effective for meta-task-oriented tasks, and model pre-training is recommended when valuable information can be extracted from unlabeled base data. Knowledge distillation works well when different models operate on different data parts. Common knowledge sharing technique aids in capturing shared patterns (data or model) across both parts, thus promoting knowledge transfer.

For tasks centered on balancing the performance of both high- and low-resource parts, data reweighting and resampling, synthetic data generation, and additional constraints are essential considerations. Data reweighting and resampling can rectify imbalance issues and enhance data distribution. Synthetic data generation techniques, such as SMOTE or GAN, can generate synthetic instances for the low-resource part, thus also rebalancing the data distribution. Additionally, when specific conditions exist for model training based on data characteristics, additional constraints can guide the model training to address imbalance issues.

Finally, \emph{exploring techniques from other branches} is a worthwhile endeavor. On one hand, although most techniques are categorized under the two branches---improving the low-resource part or balancing the high- and low-resource parts---the boundaries are not strict. Namely, techniques intended for one branch may be applicable to tasks in the other.
For instance, in the context of improving low-resource part performance, synthetic data generation and additional task-specific constraints, originally meant for balancing both parts, can potentially bolster the data and performance. Conversely, techniques like knowledge transfer, can potentially foster balancing both parts by transferring knowledge from high-resource to low-resource parts.
On the other hand, exploring novel techniques that are not commonly used in \name\ can also be beneficial, \eg, diffusion models \cite{rombach2022high}, data prompting \cite{liu2023pre}, foundation models \cite{bommasani2021opportunities}, \etc\ 
These techniques may bring new perspectives and potential solutions for addressing imbalance issues on graphs.

\section{Other Related Literature} \label{sec.other-literature}

In addition to the primary scenarios of \name, there are other contexts that fuse graph learning techniques and imbalance issues, such as graph fairness learning \cite{bose2019compositional}, cold-start recommendation \cite{schein2002methods}, and imbalanced road networks \cite{parsa2019real}. In this section, we present a concise overview of the tasks closely related to our discussion, including fairness learning on graphs and imbalanced learning in recommendations.

\subsection{Fairness Learning on Graphs} \label{sec.fairness-learning}

Fairness learning addresses the imbalance in input attributes linked to certain predictions due to historical prejudices in data collection. 
This imbalance, particularly crucial when dealing with sensitive attributes like gender, race, or education, can produce discrimination in model outcomes, fostering unfairness. Unlike other imbalanced learning forms, fairness learning aims to minimize prediction discrepancies between different subgroups, instead of typically focusing on performance on imbalanced data.

Methods to achieve fairness learning on graphs can be mainly categorized into two classes: adding constraints to model training and pre-processing input data. The first line of works adds a fairness constraint to the objective of the task \cite{bose2019compositional,guo2022fair,fan2021fair,dong2021individual,dai2021say,chen2022ba,burkholder2021certification,rahmattalabi2019exploring,anagnostopoulos2020spectral,fu2020fairness,abouzeid2022socially,ali2021fairness,yao2023path,fu2023fairness}, promoting independence between output and sensitive attributes. The second stream of works resorts to modify input features to create bias-free input, either by removing the influence of sensitive attributes \cite{rahman2019fairwalk,khajehnejad2022crosswalk,tsioutsiouliklis2021fairness,jalali2020information,spinelli2021fairdrop,li2021dyadic,kang2020inform,hussain2022adversarial,liu2023fair,ling2023learning,dong2022edits,wang2022improving,ma2022learning}, or by rendering the model invariant to these attributes \cite{dong2023reliant,buyl2020debayes,zhang2021multi,wang2022unbiased}. Apart for these two main categories, some other techniques are also deployed in this setting, such as post-processing output embeddings \cite{song2022guide}, partitioning training set \cite{ma2021subgroup,liu2022ud}, and considering post-hoc explanation of bias \cite{dong2022structural}. We refer interesting readers to a comprehensive survey \cite{dong2023fairness} for more details.

\subsection{Imbalanced learning in Recommendations}

Imbalanced learning in recommendations primarily addresses cold-start recommendations within user-item rating graphs, a bipartite graph \cite{schein2002methods,lika2014facing}. Distinct from \name, cold-start recommendation usually grapples with insufficient user-item interactions. It manifests when making recommendations for new users (user cold-start) or new items (item cold-start) that lack adequate interaction data to guide traditional collaborative filtering methods.
To mitigate the cold-start recommendation problem, numerous techniques have been employed, including meta-learning \cite{vartak2017meta,lee2019melu}, pre-training \cite{wei2021contrastive}, and knowledge distillation \cite{wang2021privileged}, \etc\
For an in-depth understanding of this challenge, we encourage readers to consult the related survey \cite{zhang2019deep}.

\section{Future Directions} \label{sec.future-directions}

In this section, we explore prospective avenues for advancing \name, building on our established taxonomies of problems (Section~\ref{sec.future-problem}) and techniques (Section~\ref{sec.future-technique}). Our goal is to stimulate research that tackles the challenges inherent to imbalanced graph data and ultimately improves the performance in various scenarios.

\subsection{Future Directions of Problems} \label{sec.future-problem}

The characteristics of graph data lead to various imbalance challenges in graph learning. Despite substantial research in this area, some facets remain underexplored and deserve further attention. Two pivotal dimensions---class and structure imbalance---provide a framework for future research when examined within our problem taxonomy. Furthermore, the varied nature of graph imbalance  necessitates diverse imbalance ratios to quantify the degree of imbalance, offering a potential cornerstone for ensuing investigations.

\stitle{Class imbalance.}
As indicated in Section~\ref{sec.problem-class}, much of the current research in \name\ centers on node-level imbalance, leaving edge- and graph-level tasks less explored.

Firstly, edge-level imbalance tasks, particularly for generic graphs, need more focus. Edge behaviors can vary across or even within a single graph, creating unique challenges \cite{zhu2023few}. Investigating imbalanced edge tasks like few-shot link prediction could significantly enhance the practicality of graph learning in real-world applications.

Secondly, graph-level imbalance tasks have received limited investigation and warrant further exploration. Beyond anomaly detection, there is a lack of research on more generic imbalanced graph classification \cite{wang2022imbalanced}, which is crucial in real-world applications. Moreover, considering the laborious nature of labeling in many molecular-related tasks, graph classification without supervision (\eg, zero-shot graph classification) becomes increasingly important, which might be potentially assisted by text information. 

\stitle{Structure imbalance.}
In the realm of structure imbalance, the majority of recent work has primarily focused on node-level imbalance. However, the intricate connectivity patterns present in graphs can offer various perspectives on structure imbalance across nodes.

For instance, beyond the well-researched degree imbalance among nodes, another type of imbalance relating to generalized node degree \cite{liu2023generalized}, which captures the abundance of contextual structures of a node, could also trigger performance disparities among nodes.
Switching to graph-level tasks, imbalance may also stem from graph size variations. Differences in graph sizes may introduce bias and impact outcomes \cite{liu2022size}, a factor that remains under-exploited and calls for further investigation.

\subsection{Future Directions of Techniques} \label{sec.future-technique}

Two major directions can be considered in order to investigate future directions of techniques for \name.

\stitle{Cross-branch technique exploration.}
As discussed in Section~\ref{sec.tech-choose}, the techniques of one branch can address the imbalance issues of another, emphasizing the potential for cross-branch technique exploration.
Thus, researchers are urged to consider this method to address imbalance issues.

\stitle{Novel technique exploration.}
As real-world applications spur artificial intelligence innovations, exploring new techniques becomes essential. For example, in the context of synthetic data generation for balancing high/low-resource parts, diffusion models \cite{rombach2022high} offer an alternative to GANs, given their superior generation capabilities in some contexts \cite{dhariwal2021diffusion,rombach2022high,saharia2022photorealistic}. Additionally, foundation models \cite{bommasani2021opportunities} like GPT models \cite{radford2018improving}, known for their potential in various domains, could potentially offer additional benefits in \name\ where the text information is available.

\subsection{Other Directions}

\name\ is an area where current studies often feel fragmented and disconnected. One of the major challenges is the absence of unified benchmark datasets and leaderboards for each research task. Such tools would not only facilitate a more direct evaluation of individual studies but also encourage healthy competition among researchers. Furthermore, there is a valuable opportunity to consolidate and review existing work on each specific task within this domain. This would produce comprehensive survey reports, offering a holistic view and serving as invaluable resources for both newcomers and seasoned researchers in the field.

\section{Conclusions} \label{sec.conclusions}

In this survey, we offer a comprehensive review of the literature on imbalanced learning on graphs. To provide a solid foundation, we begin by presenting a formal definition of the concept, followed by an introduction to basic terms in the field. Significantly, we develop two comprehensive taxonomies of \name, categorizing existing studies based on their problems and techniques. Our problem taxonomy identifies class and structure imbalance, further detailing research tasks within each branch concerning graph-related elements, namely nodes, edges, and graphs. Meanwhile, the technique taxonomy classifies the literature based on the type of imbalance and corresponding strategies deployed to rectify these imbalance on graphs. Finally, we outline promising future directions for both problems and techniques in \name, in order to spark further innovation and exploration in this increasingly critical domain.

\bibliographystyle{IEEEtran}
\bibliography{references}

\begin{IEEEbiography}[{\includegraphics[width=1in,height=1.25in,clip,keepaspectratio]{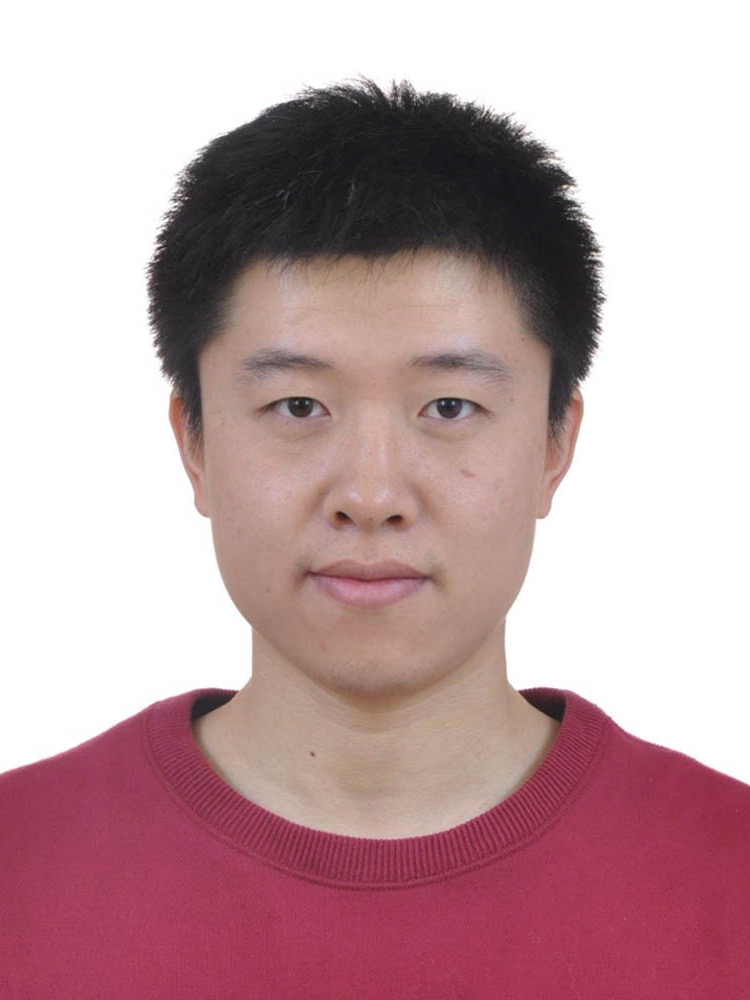}}]{Zemin Liu} 
is currently a senior research fellow with the School of Computing, National University of Singapore. 
He received his Ph.D.~degree in Computer
Science from Zhejiang University, Hangzhou, China in 2018, and B.S.~Degree in Software Engineering from Shandong University, Jinan, China in 2012.
His research interests lie in graph embedding, graph neural networks, and learning on heterogeneous information networks.
\end{IEEEbiography}

\begin{IEEEbiography}[{\includegraphics[width=1in,height=1.25in,clip,keepaspectratio]{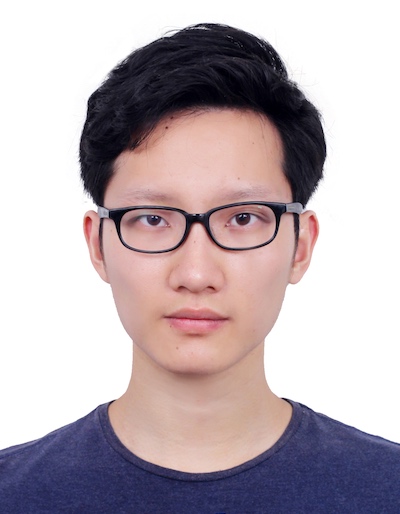}}]{Yuan Li}
is a Ph.D. student of Data Science at National University of Singapore. His current research interests include graph neural networks and learning on temporal graphs.
\end{IEEEbiography}

\begin{IEEEbiography}
[{\includegraphics[width=1in,height=1.25in,clip,keepaspectratio]{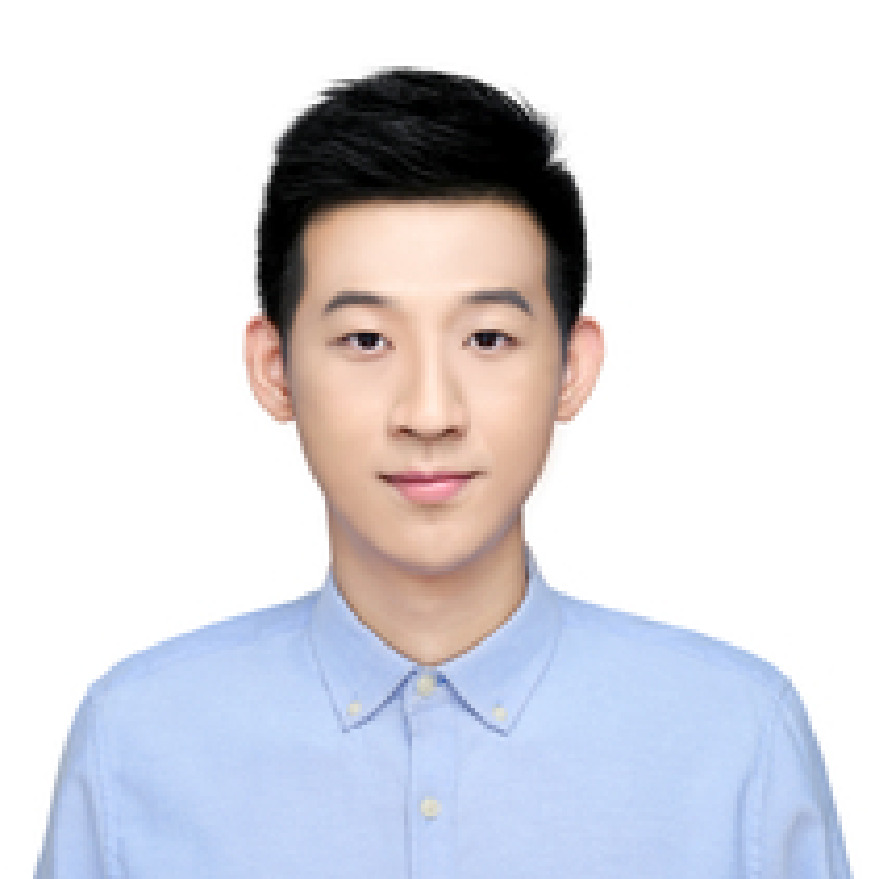}}]{Nan Chen}
is a research assistant at National University of Singapore. His current research interests include graph representation learning and equivariant machine learning.
\end{IEEEbiography}

\begin{IEEEbiography}[{\includegraphics[width=1in,height=1.25in,clip,keepaspectratio]{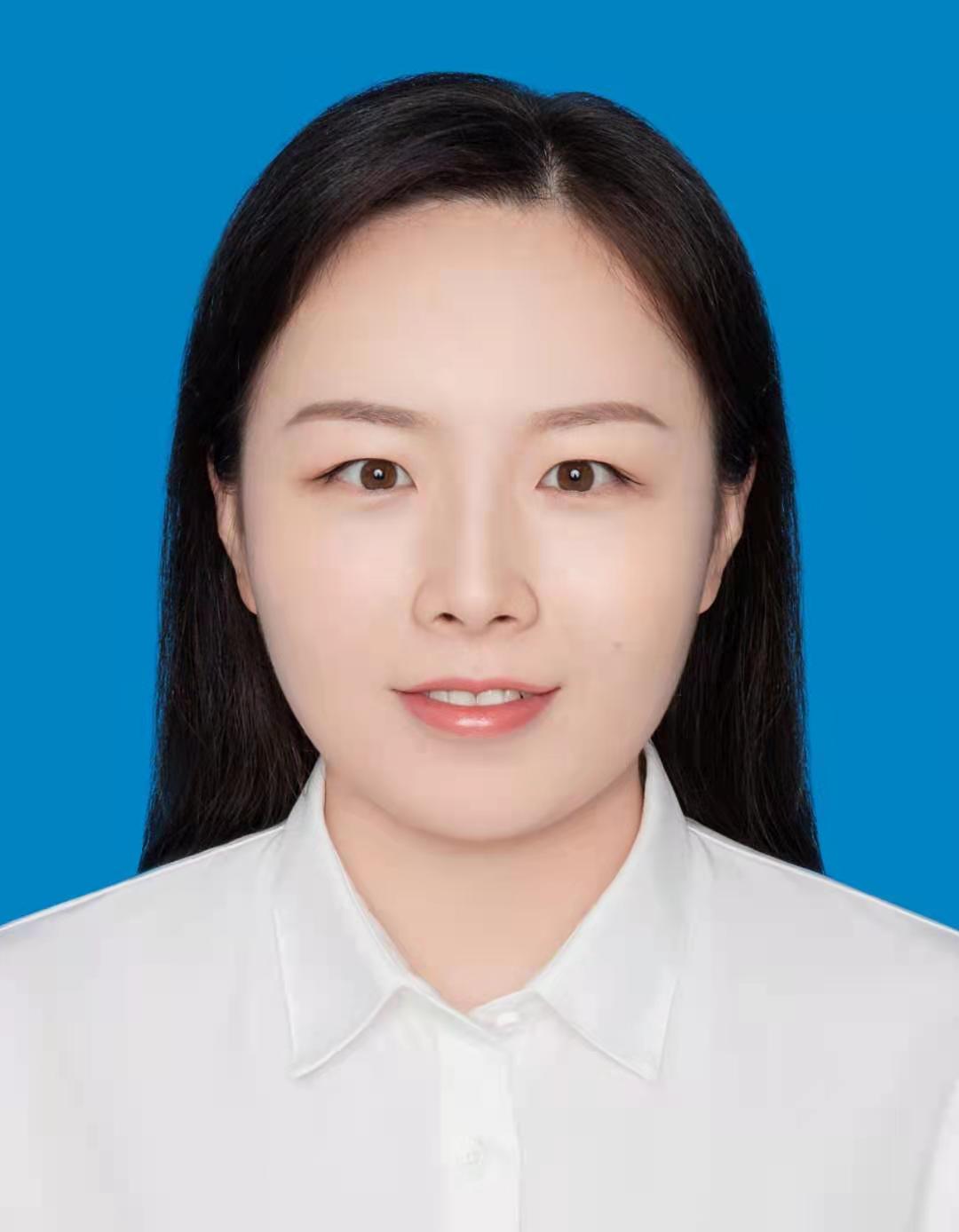}}]{Qian Wang}
is a Ph.D. student specializing in Computer Science at National University of Singapore. She has consistently demonstrated her academic excellence, having previously graduated with honors from the same institution with a Bachelor of Computing degree in Computer Science in 2023. Her current research interests lie in blockchain and graph neural networks.
\end{IEEEbiography}

\begin{IEEEbiography} [{\includegraphics[width=1in,height=1.25in,clip,keepaspectratio]{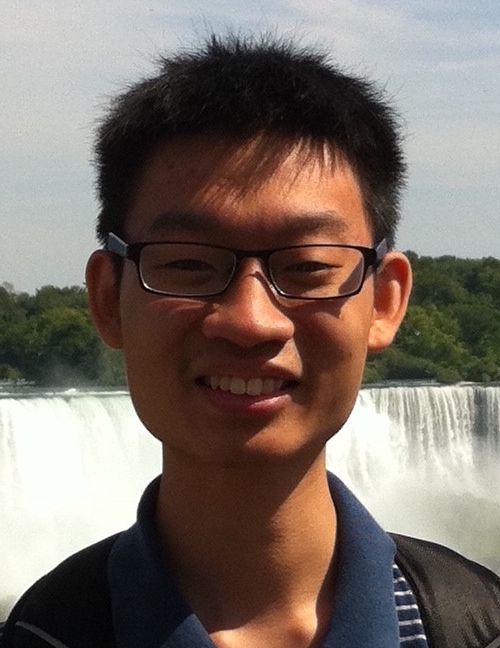}}]{Bryan Hooi} 
is an assistant professor in the
School of Computing and the Institute of Data Science in National University of Singapore. He received his Ph.D.~degree in Machine Learning from Carnegie Mellon University, USA in 2019. His research interests include methods for learning from graphs and other complex or multimodal datasets, with the goal of developing efficient and practical approaches for applications such as the detection of anomalies or malicious behavior, and automatic monitoring of medical, traffic, and environmental sensor data.
\end{IEEEbiography}

\begin{IEEEbiography}[{\includegraphics[width=1in,height=1.25in,clip,keepaspectratio]{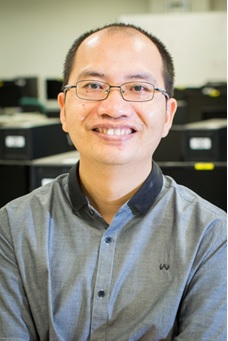}}]{Bingsheng He}
received the bachelor’s degree in computer science from Shanghai Jiao Tong University, China, in 2003, and the Ph.D. degree in computer science from the Hong Kong University of Science and Technology, Hong Kong, in 2008. He is currently a professor with the School of Computing, National University of Singapore, Singapore. His research interests are high-performance computing, distributed and parallel systems, and database systems.
\end{IEEEbiography}


\end{document}